\pdfoutput=1

\documentclass[11pt]{article}

\usepackage[final]{acl}

\usepackage{times}
\usepackage{latexsym}

\usepackage[T1]{fontenc}

\usepackage[utf8]{inputenc}

\usepackage{microtype}

\usepackage{inconsolata}

\usepackage{graphicx}

\usepackage[linesnumbered, ruled, vlined]{algorithm2e}
\usepackage{algpseudocode}
\usepackage{algorithmicx}
\usepackage{algcompatible}
\usepackage{verbatim}

\usepackage{graphicx}
\usepackage{booktabs}
\usepackage{amsmath}
\usepackage{amssymb}

\usepackage[utf8]{inputenc} 
\usepackage[T1]{fontenc}    
\usepackage{hyperref}       

\usepackage{enumerate}
\usepackage{enumitem}
\usepackage{url}            
\usepackage{booktabs}       
\usepackage{amsfonts}       
\usepackage{nicefrac}       
\usepackage{microtype}      
\usepackage{xspace}
\usepackage{cleveref}
\usepackage[table,dvipsnames]{xcolor}
\usepackage{tabularx}

\usepackage[position=bottom]{subcaption}
\usepackage{caption}

\usepackage{tabu}

\usepackage{arydshln} 

\definecolor{pos}{RGB}{0,153,0}
\definecolor{neg}{RGB}{0,0,0}
\definecolor{smpos}{RGB}{0,0,0}
\definecolor{rowx}{RGB}{242, 243, 244}
\definecolor{row}{RGB}{235, 245, 251}
\definecolor{suppcolor}{RGB}{0,0,255}
\definecolor{down}{RGB}{153, 163, 164}
\definecolor{down2}{RGB}{153, 204, 205}

\usepackage{multirow}
\usepackage{pifont}
\newcommand{\cmark}{\ding{51}}%
\newcommand{\xmark}{\ding{55}}%

\newcommand{\fref}[1]{Fig.~\ref{#1}}
\newcommand{\tref}[1]{Table~\ref{#1}}
\newcommand{\sref}[1]{Sec.~\ref{#1}}

\newcommand{\ours}{LVNet}
\newcommand{\gpto}{GPT-4o}

\newcommand{\clipb}{CLIP-B/16}
\newcommand{\resneteight}{ResNet-18}

\def\eg{\emph{e.g.~}} 

\def\ie{\emph{i.e.~}}

\usepackage{colortbl}
\definecolor{Gray}{gray}{0.90}

\usepackage{mathtools}

\def\modelname{LVNet\xspace}
\def\hksfull{Hierarchical Keyframe Selector\xspace}
\def\hks{HKS\xspace}

\def\hksAfull{Temporal Scene Clustering\xspace}
\def\hksA{TSC\xspace}

\usepackage{array}
\newcommand{\PreserveBackslash}[1]{\let\temp=\\#1\let\\=\temp}
\newcolumntype{C}[1]{>{\PreserveBackslash\centering}p{#1}}
\newcolumntype{R}[1]{>{\PreserveBackslash\raggedleft}p{#1}}
\newcolumntype{L}[1]{>{\PreserveBackslash\raggedright}p{#1}}

\newcommand{\egoschema}{EgoSchema}
\newcommand{\videoagent}{VideoAgent}

\newcommand{\videotree}{VideoTree}

\newcommand{\nextqa}{NExT-QA}
\newcommand{\intentqa}{IntentQA}
\newcommand{\traveler}{TraveLER}

\makeatletter
\newcommand{\removelatexerror}{\let\@latex@error\@gobble}
\makeatother

\title{\textit{Too Many Frames, Not All Useful}:\\  
Efficient Strategies for Long-Form Video QA}

\author{%
    {\bfseries
      Jongwoo Park $^1$ \quad
      Kanchana Ranasinghe $^1$ \quad
      Kumara Kahatapitiya $^1$ 
    }\\[0.2em]
    {\bfseries
      Wonjeong Ryu$^2$ \quad
      Donghyun Kim $^3$ \quad
      Michael S. Ryoo $^1$
    }\\[0.5em] %
      $^1$Stony Brook University \quad \quad
      $^2$KAIST AI \quad \quad
      $^3$Korea University
      \vspace{0.2em} \\
      \small{\texttt{jongwopark@cs.stonybrook.edu}} \\
      \vspace{0.8em} 
}

\begin{document}
\maketitle
\renewcommand{\arraystretch}{1.0} %

\begin{abstract}
Long-form videos that span across wide temporal intervals are highly information redundant and contain multiple distinct events or entities that are often loosely related. Therefore, when performing long-form video question answering (LVQA), all information necessary to generate a correct response can often be contained within a small subset of frames.
Recent literature leverage large language models (LLMs) in LVQA benchmarks, achieving exceptional performance, while relying on vision language models (VLMs) to convert all visual content within videos into natural language. Such VLMs often independently caption a large number of frames uniformly sampled from long videos, which is not efficient and can mostly be redundant.
Motivated by this inefficiency, we propose \modelname, a modular and training-free framework featuring a novel Hierarchical Keyframe Selector (HKS) that efficiently selects a minimal set of informative frames tailored to each question.
\modelname's modularity allows easy integration with existing approaches for more efficient LVQA. 
We achieve state-of-the-art performance among similarly configured models across four benchmark LVQA datasets: EgoSchema, NExT-QA, IntentQA, VideoMME. 
The code can be found at \href{https://github.com/jongwoopark7978/LVNet}{\texttt{https://github.com/jongwoopark7978/LVNet}}
\end{abstract}

\section{Introduction}
\label{sec:intro}

Video understanding is a long-standing vision problem \citep{Aggarwal2011HumanAA} with numerous real-world applications. 
It has been traditionally studied even before the era of differentiable representation learning, with hierarchical approaches focusing on longer videos \citep{allen1994actions,ivanov2000recognition,shi2004propagation,hongeng2004video,Ryoo2006RecognitionOC}.
Today, video understanding research involving the language modality is particularly popular, with tasks such as video question answering (QnA) that involve generating human-style conversations using large language models (LLMs) 
\citep{dataset_tapaswi2016movieqa,videoqaaaai2017,xu2017msvdqavideo}.

\begin{figure}[t]
\centering
\includegraphics[width=0.95\linewidth]{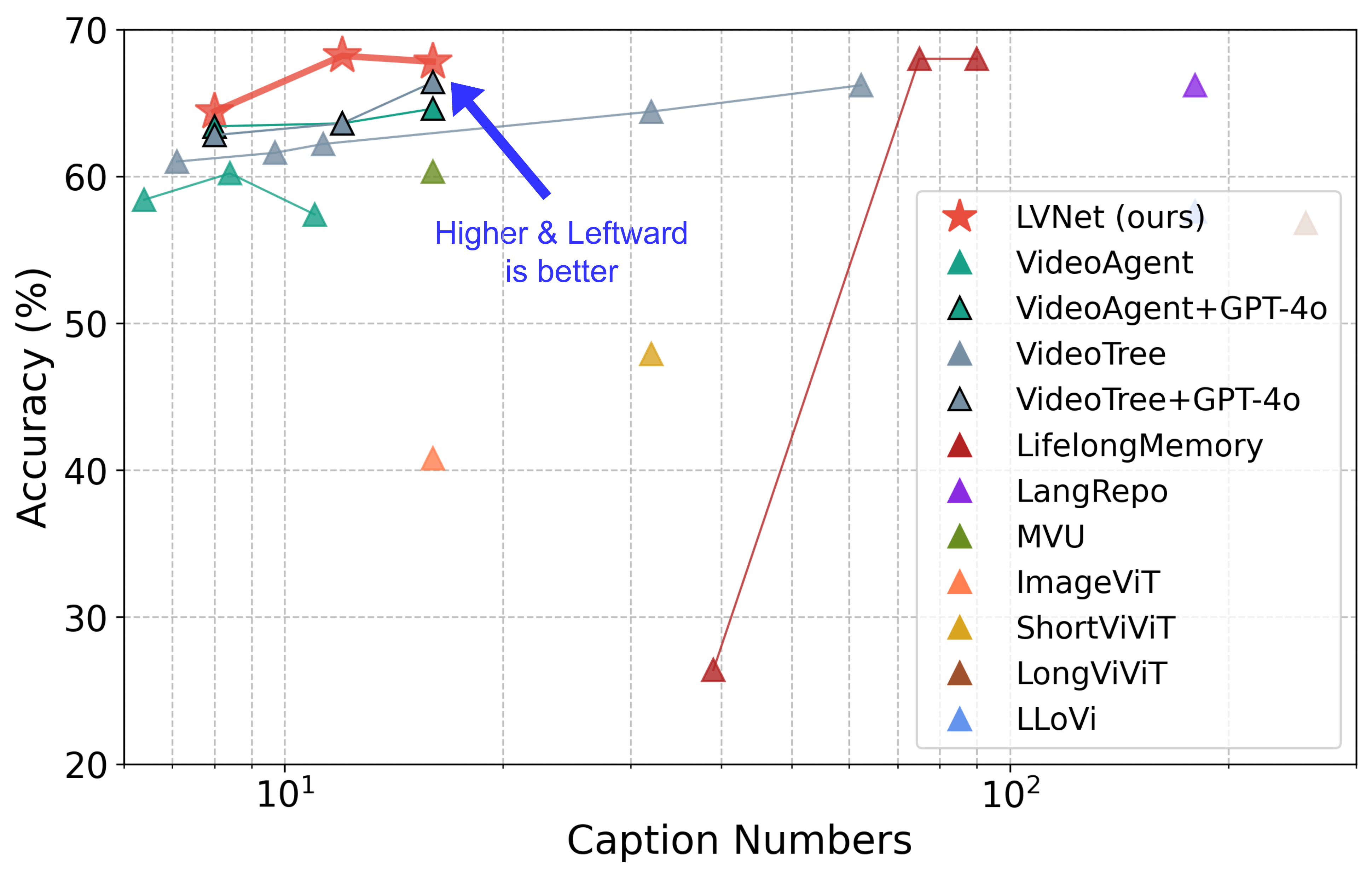}
\vspace{-0.5em}
\caption{
\textbf{High Accuracy with Low Compute:}
\modelname achieves state-of-the-art performance on EgoSchema (subset) while processing only a fraction of frame captions (below 12 per video) with the expensive LLM. More detailed analysis presented in \Cref{abl:numcaptionsaccuracy}.
}
\label{fig:teaser}
\vspace{-1.0em}
\end{figure}

Motivated by successful image-language models \citep{liu2023visual,dai2023instructblip}, several works build video LLMs \citep{Yu2023SelfChainedIM,papalampidi2023simple,Maaz2023VideoChatGPT,Wang2024Qwen2VLEV} that directly process video frames in one stage to perform QnA. However, for long videos (containing 1000s of frames) these models require processing large visual token sequences with the LLM, making inference computationally expensive or even infeasible (see \Cref{tbl:oom}).
An alternate line of works \citep{zhang2023llovi,wang2023vamos,rana2024mvu} follow multi-stage pipelines that first extract frame-level information followed by temporal modeling with an LLM. While these multi-stage works similarly encounter compute bottlenecks with increased frame processing for longer videos, it is possible to feed the LLM with descriptors of non-uniformly sampled frames \cite{Kahatapitiya2024langrepo,Wang2024VideoAgentLV,wang2024lifelongmemory}, motivating our exploration into \textit{keyframe selection}, i.e. identifying a minimal set of frames most useful for correctly answering a given video-question pair.

\begin{figure}[t]
     \centering
     \includegraphics[width=0.95\linewidth]{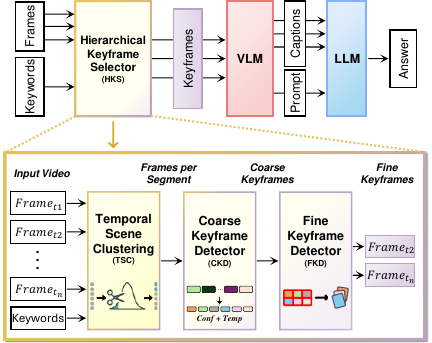}
    \vspace{-0.5em}
    \caption{
    (top) \textbf{Overview:}
    \modelname uses a \hksfull (HKS) module to select keyframes, followed by VLM \& LLM for caption and answer generation. 
    (below) \textbf{HKS Module} 
    processes dense frames with lighter modules and progressively exploits heavier, more performance-oriented modules on smaller subsets of frames to ensure efficient computation. 
    }
    \label{fig:overview}
    \vspace{-1.0em}
\end{figure}

Therein, we propose \modelname, a framework containing a novel \hksfull (\hks) that performs efficient key-frame selection followed by VLM for caption generation and LLM for answer generation as illustrated in \Cref{fig:overview}. Aligned with prior work \citep{zhang2023llovi,wang2024lifelongmemory,Wang2024VideoAgentLV}, per-frame captions are processed with a powerful LLM to generate correct answers for a given video-question pair. 
As shown in \fref{fig:teaser}, \modelname~achieves strong performance efficiently, processing only a fraction of frame descriptors (captions) with the LLM. 
Compared to feeding all captions or all frames directly to a powerful model (e.g. \gpto), our \modelname performs inference at a fraction of the cost (see \Cref{tbl:cost}).

\noindent We summarize our key contributions as follows:
\begin{enumerate}
    [label=(\alph*), leftmargin=1.5em,noitemsep,topsep=-0.4em,itemsep=-1.0ex,partopsep=0ex,parsep=1ex]
    \item \textbf{Efficient Frame Selection:} Hierarchical Keyframe Selector (HKS) efficiently extracts keyframes from sequences up to 1800 frames (hour long videos) with a filter rate over 98\%.
    \item \textbf{Video Training Free:} Our framework requires no video level training and simply uses existing off-the-shelf modules.  
    \item \textbf{Versatile Framework:} Existing methods can easily be integrated with \modelname to further boost their performance (details in \Cref{app:integrations}).
\end{enumerate}
\vspace{0.5em}

\noindent
Proposed \modelname achieves state-of-the-art results (at common inference compute) on multiple long-form video question answering benchmarks demonstrating strong performance and generality.

\section{Related Work}
\label{sec:related}

\noindent
\textbf{Video Question Answering}:
Visual question answering (VQA) involves generating open-ended textual content conditioned on an image and natural language query \citep{Agrawal2015VQAVQ}. Its video variant, Video-VQA \citep{Yu2019ActivityNetQAAD} replaces images with videos.
Multiple early datasets focus on querying objects or events based on referential and spatial relations \citep{xu2017msvdqavideo,videoqaaaai2017,Yu2019ActivityNetQAAD}. Later tasks require explicit temporal understanding of sequential events \citep{dataset_lei2018tvqa,lei-etal-2020-tvqaplus,dataset_anetqa}. 
More recent datasets focus on longer videos containing multiple actions and scenes spread over wide time intervals (termed long-form videos) \citep{dataset_xiao2021nextqa,dataset_li2022causalvidqa}.
Referred to as long-form video question answering (LVQA), these benchmarks are constructed to specifically test strong causal and temporal reasoning \citep{dataset_xiao2021nextqa} over long temporal windows \citep{Mangalam2023EgoSchemaAD}. 
Some works tackling such video VQA tasks leverage graph networks to model cross object / event relations \citep{lmappce1_hosseini-etal-2022-knowledge,model_xiao2021hgqa,model_xiao2022vgt}.
A more recent line of works integrate LLMs to tackle this task \citep{zhang2023llovi,wang2023vamos,Kahatapitiya2024langrepo,Wang2024VideoAgentLV,rana2024mvu,wang2024lifelongmemory, fan2024videoagent-M} utilizing the strong reasoning skills of LLMs. A common aspect is the use of a vision language model (VLM) to convert frame level visual information into natural language. This in turn is input to the LLM which makes a final prediction. 

Unlike these methods, \modelname incorporates a unique \hksfull that progressively reduces the number of keyframe candidates. Lighter modules are applied to dense frames, while heavier, more performance-focused modules are applied to a small subset of filtered frames. Additionally, 
\modelname does not require video-level training, unlike earlier supervised approaches.

\begin{table}[t]
    \centering
    \small
    \def\arraystretch{1.1}  %
    \setlength\tabcolsep{0.6em}  %
    \newcommand{\ourscell}{\cellcolor{row}}  %
    \scalebox{0.80}{
    \begin{tabular}{lcccccc}
        \toprule
        Feature                  & \ourscell Ours  & VA & Tr. & VT & VC & FV \\  \midrule
        \textit{(effective selection)} \\
        Uses non-uniform sampling        & \ourscell \cmark & \cmark & \cmark & \cmark & \cmark & \cmark \\
        Scene continuity-based selection & \ourscell \cmark & \xmark & \xmark & \cmark & \cmark & \cmark \\
        Robust to initial frames         & \ourscell \cmark & \xmark & \cmark & \cmark & \cmark & \cmark \\  
        Fine-grained visual refinement   & \ourscell \cmark & \xmark & \xmark & \xmark & \xmark & \xmark \\  
        \midrule
        \textit{(compute efficient)} \\
        Lightweight feature extraction  & \ourscell \cmark & \xmark & \xmark & \xmark & \cmark & \cmark \\  
        Single pass inference           & \ourscell \cmark & \xmark & \xmark & \xmark & \cmark & \cmark \\  
        Video Training Free             & \ourscell \cmark & \cmark & \cmark & \xmark & \xmark & \xmark \\  
        \bottomrule
    \end{tabular}
    }
    \vspace{-0.5em}
    \caption{\small
    LVNet exhibits unique features compared to prior work VideoAgent (VA) \cite{fan2024videoagent-M}, Traveler (Tr.) \cite{shang2024traveler}, VideoTree (VT) \cite{wang2024videotree}, VideoChat-T (VC) \cite{zeng2024timesuite}, and Frame-Voyager (FV) \cite{yu2024framevoyager} . See \Cref{app:other_work} for details.
    }
    \vspace{-0.5em}
    \label{tab:keyframe_comparison}
\end{table}

\vspace{0.5em}
\noindent
\textbf{Frame Selection in Videos}: 
The task of frame selection in videos has been long explored in video \citep{davis1997mei,Zhao_2017_ICCV} with more recent works focused directly on long-form video question answering \citep{Buch2022RevisitingT,wang2024videotree,fan2024videoagent-M,zeng2024timesuite,yu2024framevoyager}.
Most similar to our work is \citet{Wang2024VideoAgentLV} which employs an LLM based strategy for video frame selection. 
However, our \modelname differs with several unique features as summarized in \tref{tab:keyframe_comparison}.

\section{Method}
\label{sec:method}

In this section, we present our training-free (\ie zero-shot) framework for long-form video QA, \modelname. Videos are a dense form of data with even a few seconds long clip being composed of 100s of frames (individual images). In the case of long-form videos, this frame count is even greater. However, the information necessary to answer a given question is often contained in a handful of those frames. Our framework tackles this challenge of selecting an optimal and minimal set of informative frames. We refer to this as keyframe selection. Given such a set of useful frames, we also establish optimal strategies for extracting their information using modern large language models (LLMs), taking into account their sequential nature.

Our proposed \modelname comprises of three components: a \hksfull (\hks), a Vision Language Model (VLM), and a Large Language Model (LLM) as illustrated in \Cref{fig:overview}. The \hks, an efficient, hierarchical keyframe selector, is the core contribution of our work. First, the model processes 900 uniformly sampled frames and clusters them into distinct scenes
Next, it extracts keywords from a given natural language query via LLM and selects the frames most relevant to those keywords.
Finally, the selected frames are described in natural language by a more powerful and computationally intensive VLM. Finally, an LLM processes the language descriptions of the selected frames to answer a given query.

\subsection{Background}
Recent approaches utilizing LLMs for long video question answering (LVQA) \citep{zhang2023llovi,wang2023vamos,Kahatapitiya2024langrepo,rana2024mvu,Wang2024VideoAgentLV} can be viewed as a composition of three sequential stages: a) frame selection, b) VLM based frame captioning, and c) LLM based answer generation. Note that the complexity of each stage varies across methods given their focus on different aspects of the LVQA task (\eg frame selection in some is simply uniform sampling). In our work, we also follow this structure, but we focus on improving the frame selection stage. Under such a framework, our proposed \hks can serve as plug-in modules to replace the \textit{frame selection} stage and the later two stages are similar to these prior works. 

\begin{figure*}[t]
\centering
\includegraphics[width=1\linewidth]{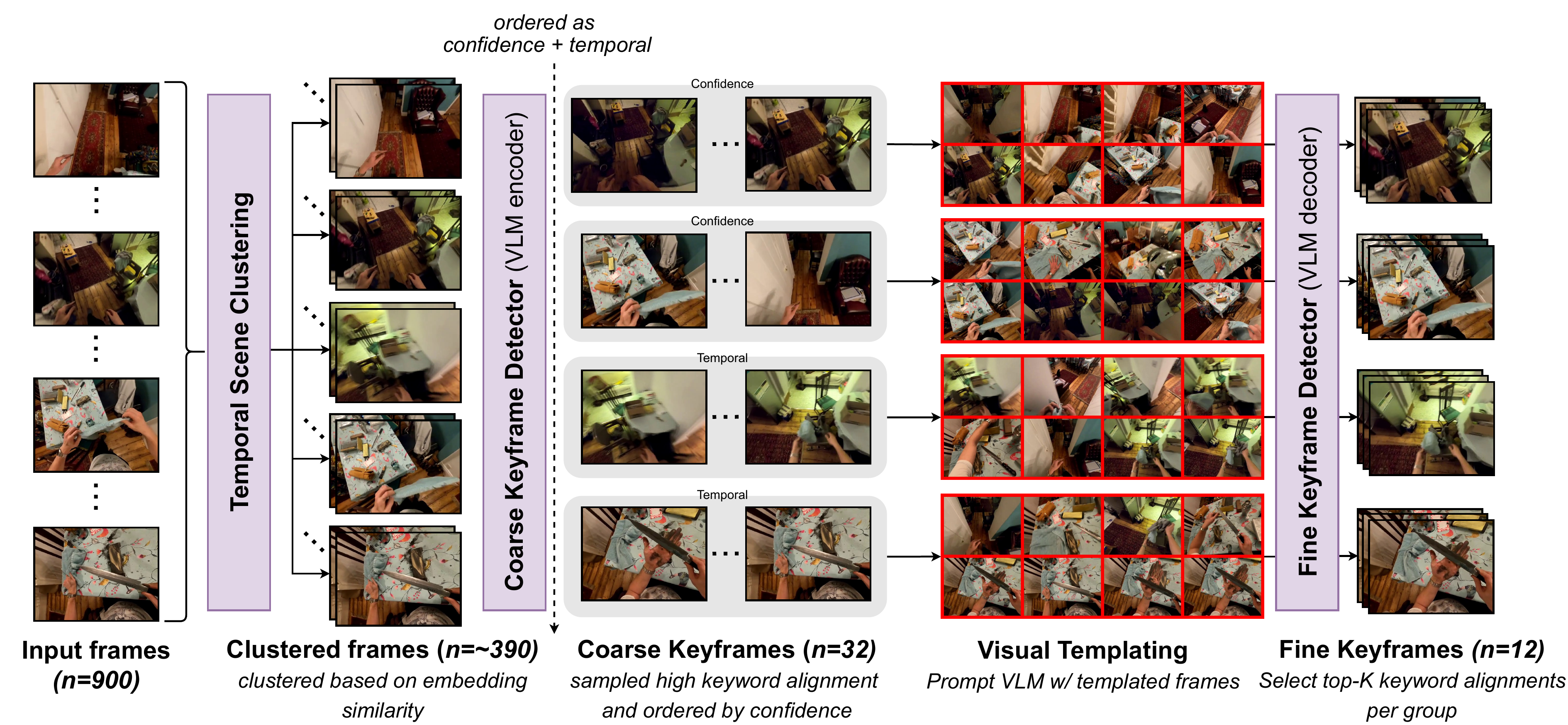}
\vspace{-1.5em}
\caption{\textbf{Qualitative example}: We illustrate a challenging long-video QA scenario from EgoSchema \citep{Mangalam2023EgoSchemaAD}. We consider an input of 900 frames, which first get clustered into scenes and subsampled to retain around 390 frames. Next, the Coarse Keyframe Detector selects only 32 frames out of them, based on the alignment with keywords (Here, keywords are extracted based on answer options, via an LLM). Such coarse keyframes are then ranked based on the combination of confidence value and temporal span, and grouped into four sets, each containing eight frames. These sets are then processed through visual templating (\ie simple concatenation across space) and fed into a VLM for Fine Keyframe Detection, resulting in just 12 frames.
}\label{fig:vis}
\vspace{-0.5em}
\end{figure*}

\subsection{Architecture}
Consider a video, $\mathbf{x} \in \mathbb{R}^{T\times C\times H\times W}$ with T, C, H, W for frames, channels, height, width respectively and its paired natural language query $\mathbf{q}$. Also consider a frame in $\mathbf{x}$ at timestamp $t$ as $\mathbf{x}[t] \in \mathbb{R}^{C\times H\times W}$. Our goal is to output a response, referred as $\mathbf{r}$, suitable for the given query $\mathbf{q}$ based on information contained in the video $\mathbf{x}$. 

Our \modelname processes a given video-query ($\mathbf{x}, \mathbf{q}$) pair to output a response, $\mathbf{\hat{r}}$. 
The \hks module initially processes this video-query pair, selects $T'$ keyframes, and outputs a deterministically sub-sampled video $\mathbf{x'} \in \mathbb{R}^{T'\times C\times H\times W}$. Each of these $T'$ frames is then passed through the captioning stage of our VLM to generate a set of natural language descriptions, 
$D = \{d_1, d_2, ...d_{T'} \}$ where $d_i$ describes the frame $\mathbf{x'}[i]$. Finally, the LLM processes all descriptions $D$ and the query $\mathbf{q}$ to generate response $\mathbf{\hat{r}}$. We illustrate this overall architecture in \Cref{fig:overview}.

\subsection{\hksfull}
We now describe our proposed \hksfull (\hks) module. As illustrated in \Cref{fig:overview}, our proposed \hks comprises of three sequential submodules, each reducing the frame count to $T_a$, $T_b$, and $T_c = T'$ respectively.

\vspace{0.5em}
\noindent
\textbf{\hksAfull (\hksA):}
The role of \hksA is to perform visual content aware preliminary frame sampling. The established approach for preliminary frame selection is uniform sampling (limited to at most 200 frames). In contrast, \hksA processes 900 to 1800 uniformly sampled frames to extract per-frame visual features using a lightweight deep neural network (\resneteight) followed by a clustering procedure to identify $n$ non-overlapping frame sets. Within each of the n sets, we uniformly sample $\leq \tau$ frames obtaining a total of $T_a \leq \tau \times n$. 
Our iterative clustering procedure is outlined in \Cref{alg:tsc}. It calculates pairwise distances between all frames accounting for intra-frame local information using the extracted per-frame features, followed by n iterative frame similarity based clustering operations. A single cluster could contain just one frame or significantly more based on frame feature similarities, leading to a \textit{non-uniform sampling of frames} across the entire video. 
This allows more frames to be sampled from the information heavy temporal regions of videos.

\vspace{0.5em}
\noindent
\textbf{Coarse Keyframe Detector (CKD):}
Unlike \hksA in the prior stage, CKD reasons across both visual and language modalities (using the paired textual query, $q$) to further sub-sample $T_a$ into $T_b$ frames. 
CKD contains three elements: keyword generation strategy, dual-encoder image-text model, and similarity based confidence assignment algorithm. Keyword generation utilizes the given query, $\textbf{q}$, alongside hand-crafted templating operations or an LLM to select or generate suitable keywords. 
The dual-encoder image-text model uses a spatially aware contrastive language image pre-training (CLIP) network from \citep{Ranasinghe2022PerceptualGI}. For confidence assignment, we construct an algorithm as outlined in \Cref{alg:ckd} which processes two lists, one of frames and one of keywords, and then calculates their pairwise likelihood of occurrence to assign each frame a confidence value (that reflects its usefulness to answer the query, $\mathbf{q}$). See ~\Cref{app:hksalgorithm} for more details.

For a single query, there can be multiple regions in a video that are highly informative but not useful or relevant in answering that query. A single query can also contain multiple different concepts and attributes that must be given attention to construct a correct answer: the keyword generation attempts to capture each of these distinct attributes. On the visual modality, a single frame will also encode multiple concepts and attributes.
Our design choice for the spatially aware CLIPpy dual-encoder VLM from \citep{Ranasinghe2022PerceptualGI} is motivated by this nature of individual frames. 
Finally, confidence assignment takes into account these multiple modes of information within each frame and the query to suitably assign confidence scores to each frame that reflects its query relevance. 
We also highlight how the confidence scores are directly linked to the related keyword (i.e. reason that makes the frame relevant), leading to better interpretability and the ability to perform further keyword-based refinement in later stages. 

\vspace{0.5em}
\noindent
\textbf{Fine Keyframe Detector (FKD):}
In the prior CKD stage, cross-modal selection utilizes a dual-encoder VLM that is constrained by the set of keywords provided and performs limited reasoning at frame level. 
In contrast, FKD uses a \textit{visual templating module} to combine multiple frames and uses VLM to generate open-ended natural language output through higher-level reasoning. The input in this stage is the set of $F_b$ frames, with each frame having an assigned confidence score and keyword. 

Our visual templating module partitions the $T_b$ frames into sets of 8 ordered by their confidence scores, arranges frame sets as grids to form a collage-style image, and annotates that image with visually identifiable tags corresponding to each frame. We further illustrate this process in \Cref{fig:vis} (see Visual Templating column). Each of these visual templated images also contain a subset of keywords that correspond to their 8 images. 
These resulting visual templated images along with a prompt containing their associated keywords and instructions to select a frame subset based on valid association between keywords and images (see \Cref{app:prompt_fkd} for details) are input to the VLM. The output of the VLM is used to select a subset of each 8 image group. These frames are collected as output of the FKD stage, overall resulting in $T_c$ frames. 

The purpose of the initial visual templating module is to allow reasoning across a set of frames using the image-text VLM (which is trained to process a single image at time). This partitioning of the input $T_b$ frames is performed based on confidence scores from the prior stage and timestamps. The eight frames with top confidence scores are grouped into the first visual template, followed by the next eight and so forth. This ensures the VLM selects both high confidence concepts and low confidence concepts, accounting for biases and weaknesses in our CKD stage. After that, we temporally reorder some image sets with low confidence scores to cover keyframes distributed across long-range segments, while the sets with high confidence scores concentrate on keyframes in short-range segments. A total of 16 low-score frames are temporally reordered in this process. The algorithm is described in \Cref{alg:fkd} and the prompting technique is explained in \Cref{app:prompt_fkd}. Our intuition is that such a mechanism allows one to best utilize the complementary strengths of two different VLMs from CKD and FKD stages for better frame selection overall.

\section{Experiments}
\label{sec:exp}

In this section, we first discuss our experimental setup followed by quantitative evaluations comparing to existing baselines and ablations of our proposed components. We then discuss the results on efficiency and scalability, and finally outline limitations.

\begin{table*}[ht!]
\centering
\def\arraystretch{1.1}  %
\setlength\tabcolsep{0.9em}  %
\resizebox{0.80\textwidth}{!}{
\begin{tabular}{lcc cc cc c}
\toprule
\multirow{2}{*}{Model} & \multicolumn{2}{c}{EgoSchema} & \multicolumn{2}{c}{NExT-QA} & \multicolumn{2}{c}{IntentQA} & \multirow{2}{*}{VT Free}\\
\cmidrule(lr){2-3} \cmidrule(lr){4-5} \cmidrule(lr){6-7}
 & Cap.\(\downarrow\) & Acc.\(\uparrow\) (\%) & Cap.\(\downarrow\) & Acc.\(\uparrow\) (\%) & Cap.\(\downarrow\) & Acc.\(\uparrow\) (\%) \\ 
\midrule
\textcolor{down}{VideoLLaMA 2} \citep{cheng2024videollama} & \textcolor{down}{-} & \textcolor{down}{53.3} & - & - & - & - & no \\
\textcolor{down}{InternVideo2} \citep{wang2024internvideo2} & \textcolor{down}{-} & \textcolor{down}{60.2} & - & - & - & - & no \\
\textcolor{down}{Tarsier} \citep{wang2024tarsier} & \textcolor{down}{-} & \textcolor{down}{61.7} & \textcolor{down}{-} & \textcolor{down}{79.2} & - & - & no \\
\midrule
Vamos \citep{wang2023vamos} & - & 48.3 & - & - & - & - & yes\\
IG-VLM \citep{kim2024igvlm} & - & 59.8 & - & 68.6 & - & 65.3 & yes \\
VIOLET \citep{fu2023violet} & 5 & 19.9 & - & - & - & - & yes \\
mPLUG-Owl \citep{ye2023mplug} & 5 & 31.1 & - & - & - & - & yes \\
VideoAgent \citep{Wang2024VideoAgentLV} & 8.4 & 54.1 & 8.2 & 71.3 & - & - & yes \\
MVU \citep{rana2024mvu} & 16 & 37.6 & 16 & 55.2 & - & - & yes \\
MoReVQA \citep{min2024morevqa} & 30 & 51.7 & 16 & 69.2 & - & - & yes \\
VFC \citep{momeni2023vfc} & - & - & 32 & 51.5 & - & -  & yes\\
ProViQ \citep{choudhury2023proviq} & 50 & 57.1 & 50 & 64.6 & - & - & yes \\
\midrule
\textcolor{down2}{VideoTree} \citep{wang2024videotree} & \textcolor{down2}{62.4} & \textcolor{down2}{61.1} & \textcolor{down2}{(56)} & \textcolor{down2}{73.5} & \textcolor{down2}{(56)} & \textcolor{down2}{66.9} & 
yes\\
\textcolor{down2}{FrozenBiLM} \citep{yang2022frozenblim} & \textcolor{down2}{90} & \textcolor{down2}{26.9} & \textcolor{down2}{-} & \textcolor{down2}{-} & \textcolor{down2}{-} & \textcolor{down2}{-} & 
yes\\

\textcolor{down2}{LifelongMemory} \citep{wang2024lifelongmemory} & \textcolor{down2}{90} & \textcolor{down2}{62.1} & \textcolor{down2}{-} & \textcolor{down2}{-} & \textcolor{down2}{-} & \textcolor{down2}{-} & 
yes\\
\textcolor{down2}{TraveLER} \citep{shang2024traveler} & \textcolor{down2}{(101)} & \textcolor{down2}{53.3} & \textcolor{down2}{(65)} & \textcolor{down2}{68.2} & \textcolor{down2}{-} & \textcolor{down2}{-} & 
yes \\
\textcolor{down2}{LangRepo} \citep{Kahatapitiya2024langrepo} & \textcolor{down2}{180} & \textcolor{down2}{41.2} & \textcolor{down2}{90} & \textcolor{down2}{60.9} & \textcolor{down2}{90} & \textcolor{down2}{59.1} & 
yes \\
\textcolor{down2}{LLoVi} \citep{zhang2023llovi} & \textcolor{down2}{180} & \textcolor{down2}{50.3} & \textcolor{down2}{90} & \textcolor{down2}{67.7} & \textcolor{down2}{90} & \textcolor{down2}{64.0} & 
yes \\
\midrule
\rowcolor{row}\modelname~(ours) & 12 & 61.1 & 12 & 72.9 & 12 & 71.7 & 
yes \\
\bottomrule
\end{tabular}
}
\vspace{-0.5em}
\caption{\textbf{Long Video Evalation:}
\modelname~achieves state-of-the-art accuracies of 61.1\%, 72.9\%, and 71.7\% on EgoSchema, NExT-QA, and IntentQA datasets respectively using just 12 frames compared to models using a similar number of captions.
Models are ordered based on number of captions processed per video. 
Models with video-level training (VT) or utilizing significantly more captions than 12 frames used by \modelname~are \textcolor{down}{de-emphasized in grey} or \textcolor{down2}{downplayed in light green} to ensure fair comparison. Numbers in parentheses () indicate the maximum number of frames used. See \sref{app:extended_res} in appendix for detailed results.  
}

\vspace{-0.5em}
\label{tab:threedatasets_brief}
\end{table*}

\begin{table}[t]
\centering
\resizebox{1.0\columnwidth}{!}{
\begin{tabular}{l c c c c}
\toprule
\textbf{Method} & \textbf{LLM Param/Type} & \textbf{VT Free} & \textbf{TS Cap.\(\downarrow\)} & \textbf{Acc.\(\uparrow\)} \\
\midrule
VideoChat-T              & 7B/OS        & no   & N/A  & 43.8 \\
Frame-Voyager           & 34B/OS       & no   & N/A  & 51.2 \\
\rowcolor{row}LVNet \small{(DS-V3)} & 37B/OS       & yes  & 24   & \textbf{53.1} \\ \midrule
VideoAgent+\small{\gpto}               & <1.8T/PP  & yes  & 24   & 51.3 \\
VideoTree+\small{\gpto}                & <1.8T/PP  & yes  & 24   & 52.3 \\
\rowcolor{row}LVNet \small{(\gpto)} & <1.8T/PP  & yes  & 24   & \textbf{53.9} \\
\bottomrule
\end{tabular}
}
\vspace{-0.5em}
\caption{\textbf{Comparison to Keyframe Selection Methods on VideoMME:} We compare LVNet with both single-stage methods that rely on video-level training (VideoChat-T, Frame-Voyager) and two-stage methods (VideoAgent, VideoTree). For the two-stage comparison, we evaluate VideoTree and VideoAgent with \gpto{} as both the captioner and LLM under the same 24-frame caption budget for a direct head-to-head comparison. \textit{Legend: DS-V3=DeepSeek-V3, OS=open-source, PP=proprietary, VT Free=no video-level training, TS Cap.=two-stage caption numbers.}. Notably, LVNet outperforms both single-stage and two-stage methods without requiring video-level training.}
\label{tab:vidmme}
\end{table}

\subsection{Experimental Setup}
\label{subsec:exp_setup}

\noindent
\textbf{Datasets}: 
Given the training free nature of our framework, we do not utilize any video datasets for training. Datasets are used purely for evaluation. We select three benchmark video visual question answering datasets focused on long-form videos for this purpose: EgoSchema \citep{Mangalam2023EgoSchemaAD}, NExT-QA \citep{dataset_xiao2021nextqa}, and IntentQA \citep{li2023intentqa_cavir}. In addition, to further highlight the strength of our approach on longer videos, we include results on VideoMME’s long split \citep{fu2024video}. These datasets are public available and can be used freely for academic research. The first dataset, EgoSchema, consists of 5031 questions and each video lasts three-minute and have multiple choice question. The second dataset, NExT-QA, is another rigorously designed video question answering benchmark containing questions that require causal \& temporal action reasoning, and common scene comprehension to correctly answer. These questions are further classified as Causal (Cau.), Temporal (Tem.), and Descriptive (Des.) and we evaluate on its validation set containing 4996 questions over 570 videos. The third dataset, IntentQA, is based on NExT-QA videos corresponding to temporal and causal reasoning quetions. It consists of 16k multiple-choice questions which are classified as Why?, How? or Before/After (B./A.). The fourth dataset, VideoMME, consists of very long videos—some up to one hour long, with an average duration of 44 minutes, and provides 900 Q\&A. Collectively, these benchmarks span \textbf{three orders of magnitude in duration}—from short clips (\(\sim\)44\,s on NExT‑QA/IntentQA), to minute‑scale videos (EgoSchema, \(\approx\)3\,min), to hour‑scale footage (VideoMME’s long split, 30–60\,min)—allowing us to stress‑test \modelname \emph{from seconds to hour‑long videos} under a single, training‑free evaluation protocol.

\vspace{0.5em}
\noindent \textbf{Model Choices \& Hyperparameters}: 
For the \hks module, we use the \resneteight~\citep{he2016resnet} for the TSC, \clipb~\citep{Ranasinghe2022PerceptualGI}~for the CKD and \gpto~for the FKD. We select \resneteight~and \clipb~due to their smaller models sizes---0.01B and 0.12B, respectively--which are significantly lighter compared to LLMs that are on a billion parameter scale.
In line with previous state-of-the-art work \cite{wang2024lifelongmemory,zhang2023llovi,wang2023vamos}, we use up to 4 NVIDIA RTX A5000 GPUs and GPT or DeepSeek APIs for running baselines and our setup in all evaluations. 
Also following prior work, we report results over single evaluation runs.

\subsection{Evaluation}
\label{subsec:results}

\begin{table*}[ht]\centering
    \label{res:ablate}%
    \subfloat[\textbf{Frame Caption Budget}: 
    LVNet ourperforms VideoAgent and VideoTree at all caption budgets when all methods use \gpto{} as both the frame captioner and LLM. \label{tab:ablation:numberofframecaptions}
    ]{
    \resizebox{0.4\textwidth}{!}{%
    \begin{tabular}{lccc}
    \toprule
    \textbf{Model} & \multicolumn{3}{c}{\textbf{Frame Captions}} \\
    \cmidrule(lr){2-4}
      & 8 & 12 & 16 \\
    \midrule
    VideoAgent+\small{\gpto}
      & 63.4 & 63.6 & 64.6 \\
    VideoTree+\small{\gpto}
      & 62.8 & 63.6 & 66.4 \\
    \rowcolor{row}
    \textbf{LVNet \small(\gpto)}
      & \textbf{64.4} & \textbf{68.2} & \textbf{67.8} \\
    \bottomrule
    \end{tabular}
    }}
    \hspace{1mm}
    \subfloat[\textbf{Visual Templating}: Combination of confidence-based \& temporal ordering gives the best performance. \label{tab:ablation:visualtemplateorder}
    ]{
    \resizebox{0.25\textwidth}{!}{
    \begin{tabular}{lc}
    \toprule
      \multicolumn{1}{l}{Templating Order} & Acc. $\uparrow$\\
    \midrule
     Temporal    & 65.2 \\
     Confidence  & 67.6 \\  \rowcolor{row}
     \textbf{Hybrid} (both) & \textbf{68.2} \\
    \bottomrule
    \end{tabular}
    }}
    \hspace{1mm}
    \subfloat[\textbf{HKS Ablation}: \modelname accuracy consistently improves with each HKS sub-module. \label{tab:ablation:effectofcomponents}
    ]{
    \resizebox{0.25\textwidth}{!}{
    \begin{tabular}{cccc}
    \toprule
      TSC & CKD & FKD & Acc. $\uparrow$ \\ \midrule
      \xmark & \xmark & \xmark & 62.6 \\
      \cmark & \xmark & \xmark & 64.5 \\
      \cmark & \cmark & \xmark & 65.8 \\
      \rowcolor{row} \cmark & \cmark & \cmark & \textbf{68.2} \\
    \bottomrule
    \end{tabular}
    }}
    \vspace{-2mm}
    \caption{\textbf{Ablation study} on EgoSchema \citep{Mangalam2023EgoSchemaAD}: We evaluate different design decisions of our framework on EgoSchema 500-video subset for zero-shot VQA.}
\end{table*}

\begin{table}[t]
\centering
\small
\def\arraystretch{1.3}  %
\setlength\tabcolsep{0.6em}  %
\resizebox{1.0\columnwidth}{!}{
\begin{tabular}{lcccc}
\toprule
Method      & LLM AP & Frames $\uparrow$ & LLM frames $\downarrow$ & Acc. $\uparrow$   \\ \midrule
Qwen-VL  & 7B     & 128    & 128        & 37.8     \\
Qwen-VL  & 7B     & 256    & 256        & OOM      \\
Qwen-VL  & 7B     & 1800   & 1800       & OOM      \\ \rowcolor{row}
LVNet (DS-V3)& 37B   & 1800   & 24         & 53.1     \\ \bottomrule
\end{tabular}
}
\vspace{-0.5em}
\caption{
\textbf{Single-Stage Method Comparison:}
We report Accuracy on VideoMME long split (average video length 41 mins) along with
LLM active parameters (LLM AP), total frames processed, and frames input to LLM as tokens / captions (LLM frames).
LVNet DeepSeek-V3 (DS-V3) variant is used. 
Processing lengthy frame sequences with single-stage models at fixed compute becomes infeasible. Inference tested on 4 x 24GB A5000 GPUs. Similar findings in \citet{zeng2024timesuite}.
}
\label{tbl:oom}
\end{table}

\paragraph{Quantitative Results:} We evaluate \modelname~on the \egoschema, \nextqa, and \intentqa~dataset and present our results in \tref{tab:threedatasets_brief}. Models with video-caption pretraining are \textcolor{down}{de-emphasized in grey} to ensure fairness with image-level pertaining. Models utilizing significantly more captions than the 12 frames are \textcolor{down2}{downplayed in light green} to consider caption efficiency.
We reiterate how number of captions input to LLM affects inference compute cost of methods the most.

For \egoschema (fullset), \modelname achieves 61.1\%, the highest among the models utilizing approximately 12 captions. This result outperforms \videoagent, the next best model using 8.4 captions, by +7\%, is on par with \videotree~while using only 1/5 of the captions, and outperforms \traveler~by +7.8\% while utilizing only 12\% of the captions.

We next evaluate on \nextqa~dataset, which has a particular focus on both temporal and casual reasoning based question-answer pairs. \modelname achieves state-of-the-art performance on this benchmark outperforming prior work among the models utilizing approximately 12 captions. In fact, our \modelname outperforms \videoagent~by +1.6\%.

In \intentqa~dataset, \modelname~outperforms all prior work, including \textcolor{down}{de-emphasized} models with video-caption pretraining and \textcolor{down2}{downplayed} models utilizing significantly compute (captions input to LLM) than ours. In fact, \modelname~shows a substantial improvement of +4.8\% over the next best \videotree, while using only 13\% of captions (12 vs 90).

Lastly, \tref{tab:vidmme} shows \modelname's performance on VideoMME's long split \citep{fu2024video}, which features videos up to an hour long—significantly exceeding the 12-minute average in MLVU \citep{MLVU} or the 17-minute average in the overall VideoMME. In combination with an open-source LLM (DeepSeek-V3), \modelname\ outperforms the single-stage keyframe selection methods VideoChat-T \citep{zeng2024timesuite} and Frame-Voyager \citep{yu2024framevoyager} by +9.3\% and +1.9\%, respectively, without any video-level training. Moreover, under a direct head-to-head setting where VideoAgent, VideoTree, and \modelname\ all use \gpto{} as both the frame captioner and LLM with the same 24-frame budget, \modelname\ achieves 53.9\% accuracy, outperforming VideoAgent(51.3\%) and VideoTree(52.3\%) by +2.6\% and +1.6\%, respectively, further confirming its effectiveness on very long videos. See \Cref{app:extended_videomme} for the detailed results.

\paragraph{Qualitative Examples:} Due to space, qualitative open-ended responses comparing LVNet to uniform sampling are provided in the Appendix (see \Cref{fig:qualanswer}).

\begin{table}[t]
\centering
\small
\def\arraystretch{1.3}  %
\setlength\tabcolsep{0.9em}  %
\resizebox{1.0\columnwidth}{!}{
\begin{tabular}{lccccc}
\toprule
Model  & FV $\uparrow$ & FL $\downarrow$ & VC $\downarrow$ & TC $\downarrow$ & Acc. $\uparrow$  \\ \midrule
GPT-4o & 384  & 384 & \$2.88 &  $\sim$\$2592  & 65.3 \\ \rowcolor{row}
LVNet (\gpto)  & 1800 & 24  & \$0.19 & \$171  & 53.9 \\ \bottomrule
\end{tabular}
}
\vspace{-0.5em}
\caption{
\textbf{API Comparison:}
We perform cost comparison with using \gpto~directly on frames vs with LVNet. We report accuracy on VideoMME long split (Acc), frames processed per video (FV), frames / captions input to LLM (FL), per video cost (VC), and total evaluation cost (TC). Our \modelname achieves competitive performance at over 10x less inference cost. \gpto~accuracy result from official benchmark leaderboard. 
}
\label{tbl:cost}
\end{table}

\subsection{Efficiency \& Scalability}
\label{subsec:efficiency_scalability}

\noindent\textbf{Stage-wise runtime, FPS, and memory.}
We report a detailed efficiency profile for each LVNet component on a single VideoMME-long video (RTX-6000 Ada). As shown in \Cref{tbl:lvnet_stage}, the three-stage HKS (TSC$\rightarrow$CKD$\rightarrow$FKD) contributes 36\% of the total runtime, confirming that most compute (64\%) lies in captioning and LLM Q\&A, which are common overheads to two-stage pipelines.
\begin{table}[t]
\centering
\def\arraystretch{1.3}  %
\setlength\tabcolsep{0.9em}  %
\resizebox{1.0\columnwidth}{!}{
\begin{tabular}{lccccc}
\toprule
Stages & FPS & Mem. & Frames & Lat. (s) & Lat. (\%) \\ \midrule
HKS-1/3~(TSC) & 182 & 19 & 1800 & 9.9 & 16 \\
HKS-2/3~(CKD) & 138 &  9 &  427 & 3.1 &  5 \\
HKS-3/3~(FKD) &   1 &  - &    8 & 9.2 & 15 \\
\midrule
Captioning     &   1 &  - &   24 & 38.7 & 62 \\
\midrule
LLM Q\&A       &   - &  - &   -  & 1.0 &  2 \\ \bottomrule
\end{tabular}
}
\vspace{-0.5em}
\caption{
\textbf{LVNet Stage-wise Performance:}
The table reports per-module elapsed time, FPS, and peak GPU memory for a single VideoMME-long video processed on an RTX-6000 Ada. Within LVNet, the three stage Hierarchical Keyframe Selectors contributes 36~\% of the total runtime. It indicates that most compute lies in the captioning and LLM Q\&A steps, which are overheads shared by all two-stage pipelines, rather than in the keyframe selection. Mem. and Lat. in the header stands for the GPU memory (GB) and Latency.
}
\label{tbl:lvnet_stage}
\end{table}

\noindent\textbf{Direct runtime comparison to two-stage baselines.}
Complementing the internal latency breakdown in \Cref{tbl:lvnet_stage}, we measure end-to-end per-video inference time for LVNet, VideoAgent, and VideoTree under matched conditions on a single NVIDIA RTX A5000 GPU with a 180-frame input video. As shown in \Cref{tbl:runtime_compare}, LVNet finishes in 25.6\,s (7.0 FPS), achieving 3.4$\times$ and 2.5$\times$ speedups over VideoAgent (87.3\,s, 2.1 FPS) and VideoTree (64.2\,s, 2.8 FPS), respectively. The main gap comes from early-stage iterations: VideoAgent takes 40.5\,s in the initial selection stage as it invokes a heavy LLM iteratively in every step when constructing keyframe grids. VideoTree takes 52.4\,s as it similarly runs a heavy VLM captioner and LLM repeatedly in the first stage (Adaptive Breadth Expansion) to form keyframe clusters. This early-stage, iterative use of heavy VLM/LLM modules dominates their latency, and many of the frames they process turn out to be uninformative for answering the question. On the other hand, LVNet’s lightweight HKS (TSC+CKD+FKD) takes only 6.7\,s and applies the expensive captioner+LLM only once on the final selected frames.
\begin{table}[t]
\centering
\small
\def\arraystretch{1.3}
\setlength\tabcolsep{1.0em}
\begin{tabular}{lcc}
\toprule
\textbf{Model} & \textbf{Inference Time (s)$\downarrow$} & \textbf{FPS$\uparrow$} \\
\midrule
VideoAgent & 87.3 & 2.1 \\
VideoTree  & 64.2 & 2.8 \\
\rowcolor{row}
LVNet      & \textbf{25.6} & \textbf{7.0} \\
\bottomrule
\end{tabular}
\vspace{-0.5em}
\caption{
\textbf{Direct runtime comparison to two-stage baselines.}
Wall-clock inference time on a 180-frame input video measured using a single NVIDIA RTX A5000 GPU.
LVNet achieves 3.4$\times$ and 2.5$\times$ speedups over VideoAgent and VideoTree, respectively.
}
\label{tbl:runtime_compare}
\end{table}

\noindent\textbf{End-to-end throughput and single-GPU footprint.}
\Cref{tbl:efficiency} contrasts end-to-end wall-clock speed and memory. LVNet processes 1{,}800 frames in 61.9\,s (29.1 FPS) on a single GPU using 19\,GB, while Qwen2.5-VL-72B requires 8 GPUs and 373\,GB yet runs $\sim$40$\times$ slower at 0.7 FPS. Qwen2.5-VL-72B encounters out of memory (OOM) at 24 frames. Despite architectural and hardware differences, \Cref{tbl:lvnet_stage} and  \Cref{tbl:efficiency}   show that LVNet’s keyframe selector is scalable to process large number of frames in long-form video as it is lightweight, single-GPU-friendly, and provides high FPS.
\begin{table}[t]
\centering
\small
\def\arraystretch{1.3}  %
\setlength\tabcolsep{0.9em}  %
\resizebox{1.0\columnwidth}{!}{
\begin{tabular}{lcccc}
\toprule
Model & Frames & Memory~(GB) & Inference time~(s) & FPS \\ \midrule
Qwen2.5-VL-72B & 8  & 355 & 11.8 & 0.7 \\
Qwen2.5-VL-72B & 12 & 373 & 16.8 & 0.7 \\
Qwen2.5-VL-72B & 24 & OOM & OOM  & OOM \\ \rowcolor{row}
LVNet          & 1800 & 19  & 61.9 & 29.1 \\ 
\bottomrule
\end{tabular}
}
\vspace{-0.5em}
\caption{
\textbf{Model Efficiency Comparison:}
End-to-end, LVNet handles 1{,}800 frames in 61.9~s~(29.1~FPS) while using just 19~GB on a single GPU, whereas Qwen2.5-VL-72B requires eight GPUs and 373~GB yet runs $\approx 40\times$ slower at 0.7~FPS. Despite architectural and hardware differences, these figures show that LVNet keyframe selector is lightweight and that the overall system delivers high FPS while remaining single-GPU friendly for long-form video.
}
\label{tbl:efficiency}
\end{table}

\noindent\textbf{Single-stage scaling limits on long videos.}
Table~\ref{tbl:oom} shows that single-stage models (e.g., Qwen-VL \citep{bai2023versatilevisionlanguagemodel-qwenvl}) run OOM beyond 128 frames, whereas LVNet achieves higher accuracy with only 24 keyframes passed to the LLM, maintaining tractable compute. It shows the LVNet's scalability of processing large number of frames with high accuracy. 

\noindent\textbf{API cost comparison.}
Table~\ref{tbl:cost} compares inference cost and performance against purely GPT-4o prompting, showing that \modelname delivers competitive accuracy with over 10$\times$ lower per-video LLM cost, thanks to efficient keyframe selection.

\noindent\textbf{Scalability across frame budget and LLM upgrades.}
\Cref{tbl:frames_llm} summarizes two complementary scaling axes. \textit{Left:} on VideoMME‑Long, increasing the frame budget from 12 to 24 yields a +3.8\% gain. Conversely, on EgoSchema’s 3-min videos, the 12-frame setting is optimal (68.2 \%), indicating extra keyframes help LVNet for longer videos. It is consistent with our keyframe‑driven design that benefits from additional informative frames when videos are very long. \textit{Right:} LVNet improves monotonically with stronger LLMs (GPT‑3.5 $\rightarrow$ GPT‑4 $\rightarrow$ GPT‑4o) without any retraining, highlighting the modularity and scalability of LVNet with newer LLMs.
\begin{table}[t]\centering
    \label{tab:eff_scaling}\vspace{3mm}
    \subfloat[\textbf{\# Frames in VideoMME.} \label{tab:frames_acc_eff}]{
        \resizebox{0.5\columnwidth}{!}{
        \begin{tabular}{cc}
        \toprule
        Frames & Acc. (\%) \\
        \midrule
        12 & 50.1 \\
        \rowcolor{row}
        24 & \bf{53.9 \; (+3.8)} \\
        \bottomrule
        \end{tabular}}}
    \hspace{3mm}
    \subfloat[\textbf{Choice of LLM} 
    \label{tab:llm_choice_eff}]{
        \resizebox{0.38\columnwidth}{!}{
        \begin{tabular}{lc}
        \toprule
        LLM & Acc. (\%) \\
        \midrule
        GPT\mbox{-}3.5 & 61.0 \\
        GPT\mbox{-}4   & 65.4 \\
        \rowcolor{row}
        GPT\mbox{-}4o  & \bf{68.2} \\
        \bottomrule
        \end{tabular}}}
    \vspace{-2mm}
    \caption{\textbf{Scalability of LVNet.} Left: accuracy improves as the frame budget increases for very long videos. Right: newer LLMs yield higher accuracy without retraining, reflecting LVNet’s modularity.}
\label{tbl:frames_llm}
\end{table}

\subsection{Ablations}
\label{subsec:ablate}

In this section, we present ablations on key design decisions such as the sorting order in FKD, the number of frames for captions, and the effect of different components in HKS. In all ablations, we use a subset of EgoSchema \citep{Mangalam2023EgoSchemaAD}, composed of 500 videos. Additional ablations about \textit{Choice of LLM} and  \textit{Effect of Patch Size on Keyword Matching in CKD} are in \Cref{app:additionalablation} 

\label{abl:numcaptionsaccuracy}
\paragraph{Number of Frame Captions:} We conduct an ablation on the number of frame captions, directly comparing our approach with VideoAgent \citep{Wang2024VideoAgentLV} and VideoTree \citep{wang2024videotree} under the same \gpto{} backbone for both frame captioning and LLM reasoning. As shown in \tref{tab:ablation:numberofframecaptions}, \modelname~achieves the highest accuracy of 68.2\% with 12 captions, outperforming both \videoagent{} and \videotree{} (63.6\%) by +4.6\% absolute. \modelname~also consistently outperforms the baselines at 8 captions (+1.0\% / +1.6\%) and 16 captions (+3.2\% / +1.4\%) compared to \videoagent{} / \videotree{}..

\paragraph{Visual Templating Order:} In visual templating, prioritizing frames by keyword confidence scores followed by reordering low-confidence frames based on timestamp proves more effective than using confidence scores or temporal order alone, as shown in \tref{tab:ablation:visualtemplateorder}. In this hybrid approach, high-confidence frames capture short but important segments of videos, while low-confidence keyframes, which are crucial but visually challenging for keyword matching, are temporally ordered to cover broader segments. This hybrid approach outperforms solely temporal ordering or confidence-based ordering by +3\% and +0.6\%, respectively.

\paragraph{Effect of Hierarchical Keyframe Modules:} \tref{tab:ablation:effectofcomponents} demonstrates the impact of incrementally adding the temporal scene clustering (TSC), coarse keyframe detector (CKD), and fine keyframe detector (FKD) modules. Without any of these modules, the model relies on uniform sampling and achieves 62.6\%. When TSC is added and 12 frames are selected uniformly, the accuracy increases to 64.5\%. Adding both TSC and CKD raises the accuracy to 65.8\%. Finally, incorporating all three modules---TSC, CKD, and FKD---into the model, which is \modelname, results in an accuracy of 68.2\%. This demonstrates the importance of including all modules in \modelname~for optimal performance.

\section{Conclusion}
\label{sec:conclusion}
We proposed a novel approach for Long-form Video Question Answering (LVQA) that achieves state-of-the-art performance compared to the model using the similar-scale captions across four benchmarks. Our \hksfull demonstrates the effectiveness of keyframe selection in understanding a very long-form video QA. Additionally, we highlight the zero-shot capability for long-form video comprehension of our \modelname framework, which requires no video-level training. Our experiments showcase its significant advantage over existing single-stage and two-stage keyframe selectors. 

\section*{Limitations}

Despite the effectiveness of \modelname, as demonstrated by benchmark experiments and comprehensive ablations, our study has certain limitations, which we discuss below.

\begin{itemize}[leftmargin=*]
    \setlength\itemsep{1pt}
    \item First, we acknowledge that we are unable to evaluate \modelname~and other models with all available VLMs or LLMs due to computational constraints and high costs. However, we carefully select GPT-4o and DeepSeek-v3, the LLMs commonly used in video understanding research, for our main experiments and provide ablation studies comparing various LLMs (\eg GPT-3.5, GPT-4, and GPT-4o) to ensure a fair performance comparison, as presented in \Cref{tab:ablation:numberofframecaptions} and \Cref{tab:llm_choice_eff}.
    
    \item Our hierarchical keyframe selector consists of three components: TSC, CKD, and FKD. While we demonstrated the effectiveness of each component in \Cref{tab:ablation:effectofcomponents}, we did not have the time or resources to develop a unified module that could replace all three. Although this is beyond the scope of this paper, exploring a more efficient implementation that integrates these three modules into a single model would be an interesting direction for future research.

    \item Like any LLM-based approach, \modelname~is sensitive to prompting. To ensure the transparency, we provide examples of these prompts in \Cref{fig:qualanswer} and \Cref{fig:fkd_prompt}. We also plan to release the code to enable further exploration by other researchers.

    \item Finally, we acknowledge that, as our approach is zero-shot, any inherent limitations or biases in the pretrained models may persist in the outputs of \modelname.

\end{itemize}

\paragraph{Acknowledgements:} This research was financially supported by the Ministry of Trade, Industry, and Energy (MOTIE), Korea, under the “Global Industrial Technology Cooperation Center(GITCC) program” supervised by the Korea Institute for Advancement of Technology (KIAT).(Task No. P0028420). This research was supported by the National Research Council of Science \& Technology(NST) grant by the Korea government(MSIT) (No. GTL25041-000).

\bibliography{custom}

\begin{thebibliography}{65}
\providecommand{\natexlab}[1]{#1}

\bibitem[{Aggarwal and Ryoo(2011)}]{Aggarwal2011HumanAA}
Jake~K. Aggarwal and Michael~S. Ryoo. 2011.
\newblock \href {https://api.semanticscholar.org/CorpusID:5388357} {Human activity analysis}.
\newblock \emph{ACM Computing Surveys (CSUR)}, 43:1 -- 43.

\bibitem[{Agrawal et~al.(2015)Agrawal, Lu, Antol, Mitchell, Zitnick, Parikh, and Batra}]{Agrawal2015VQAVQ}
Aishwarya Agrawal, Jiasen Lu, Stanislaw Antol, Margaret Mitchell, C.~Lawrence Zitnick, Devi Parikh, and Dhruv Batra. 2015.
\newblock \href {https://api.semanticscholar.org/CorpusID:3180429} {Vqa: Visual question answering}.
\newblock \emph{International Journal of Computer Vision}, 123:4 -- 31.

\bibitem[{Allen and Ferguson(1994)}]{allen1994actions}
James~F Allen and George Ferguson. 1994.
\newblock Actions and events in interval temporal logic.
\newblock \emph{Journal of logic and computation}, 4(5):531--579.

\bibitem[{Bai et~al.(2023)Bai, Bai, Yang, Wang, Tan, Wang, Lin, Zhou, and Zhou}]{bai2023versatilevisionlanguagemodel-qwenvl}
Jinze Bai, Shuai Bai, Shusheng Yang, Shijie Wang, Sinan Tan, Peng Wang, Junyang Lin, Chang Zhou, and Jingren Zhou. 2023.
\newblock \href {https://arxiv.org/abs/2308.12966} {Qwen-vl: A versatile vision-language model for understanding, localization, text reading, and beyond}.
\newblock \emph{Preprint}, arXiv:2308.12966.

\bibitem[{Buch et~al.(2022)Buch, Eyzaguirre, Gaidon, Wu, Fei-Fei, and Niebles}]{Buch2022RevisitingT}
S.~Buch, Cristobal Eyzaguirre, Adrien Gaidon, Jiajun Wu, Li~Fei-Fei, and Juan~Carlos Niebles. 2022.
\newblock \href {https://api.semanticscholar.org/CorpusID:249375461} {Revisiting the “video” in video-language understanding}.
\newblock \emph{2022 IEEE/CVF Conference on Computer Vision and Pattern Recognition (CVPR)}, pages 2907--2917.

\bibitem[{Chen et~al.(2024{\natexlab{a}})Chen, Xue, Li, Hu, Zhu, Li, Fang, Tang, Yang, Liu, He, Yin, Molchanov, Kautz, Fan, Zhu, Lu, and Han}]{chen2024scalinglongcontextvisual-longvila}
Yukang Chen, Fuzhao Xue, Dacheng Li, Qinghao Hu, Ligeng Zhu, Xiuyu Li, Yunhao Fang, Haotian Tang, Shang Yang, Zhijian Liu, Ethan He, Hongxu Yin, Pavlo Molchanov, Jan Kautz, Linxi Fan, Yuke Zhu, Yao Lu, and Song Han. 2024{\natexlab{a}}.
\newblock \href {https://arxiv.org/abs/2408.10188} {Longvila: Scaling long-context visual language models for long videos}.
\newblock \emph{Preprint}, arXiv:2408.10188.

\bibitem[{Chen et~al.(2024{\natexlab{b}})Chen, Wang, Cao, Liu, Gao, Cui, Zhu, Ye, Tian, Liu et~al.}]{chen2024intervl25}
Zhe Chen, Weiyun Wang, Yue Cao, Yangzhou Liu, Zhangwei Gao, Erfei Cui, Jinguo Zhu, Shenglong Ye, Hao Tian, Zhaoyang Liu, et~al. 2024{\natexlab{b}}.
\newblock Expanding performance boundaries of open-source multimodal models with model, data, and test-time scaling.
\newblock \emph{arXiv preprint arXiv:2412.05271}.

\bibitem[{Chen et~al.(2024{\natexlab{c}})Chen, Wang, Tian, Ye, Gao, Cui, Tong, Hu, Luo, Ma et~al.}]{chen2024far_internvl2}
Zhe Chen, Weiyun Wang, Hao Tian, Shenglong Ye, Zhangwei Gao, Erfei Cui, Wenwen Tong, Kongzhi Hu, Jiapeng Luo, Zheng Ma, et~al. 2024{\natexlab{c}}.
\newblock How far are we to gpt-4v? closing the gap to commercial multimodal models with open-source suites.
\newblock \emph{Science China Information Sciences}, 67(12):220101.

\bibitem[{Cheng et~al.(2024)Cheng, Leng, Zhang, Xin, Li, Chen, Zhu, Zhang, Luo, Zhao et~al.}]{cheng2024videollama}
Zesen Cheng, Sicong Leng, Hang Zhang, Yifei Xin, Xin Li, Guanzheng Chen, Yongxin Zhu, Wenqi Zhang, Ziyang Luo, Deli Zhao, et~al. 2024.
\newblock Videollama 2: Advancing spatial-temporal modeling and audio understanding in video-llms.
\newblock \emph{arXiv preprint arXiv:2406.07476}.

\bibitem[{Choudhury et~al.(2023)Choudhury, Niinuma, Kitani, and Jeni}]{choudhury2023proviq}
Rohan Choudhury, Koichiro Niinuma, Kris~M Kitani, and L{\'a}szl{\'o}~A Jeni. 2023.
\newblock Zero-shot video question answering with procedural programs.
\newblock \emph{arXiv preprint arXiv:2312.00937}.

\bibitem[{Dai et~al.(2023)Dai, Li, Li, Tiong, Zhao, Wang, Li, Fung, and Hoi}]{dai2023instructblip}
Wenliang Dai, Junnan Li, Dongxu Li, Anthony Meng~Huat Tiong, Junqi Zhao, Weisheng Wang, Boyang Li, Pascale Fung, and Steven Hoi. 2023.
\newblock Instructblip: Towards general-purpose vision-language models with instruction tuning.
\newblock \emph{arXiv preprint arXiv:2305.06500}.

\bibitem[{Davis and Bobick(1997)}]{davis1997mei}
James Davis and Aaron Bobick. 1997.
\newblock The representation and recognition of action using temporal templates.
\newblock In \emph{Proceedings of the IEEE International Conference on Computer Vision}, pages 2736--2744.

\bibitem[{Fan et~al.(2024)Fan, Ma, Wu, Du, Li, Gao, and Li}]{fan2024videoagent-M}
Yue Fan, Xiaojian Ma, Rujie Wu, Yuntao Du, Jiaqi Li, Zhi Gao, and Qing Li. 2024.
\newblock Videoagent: A memory-augmented multimodal agent for video understanding.
\newblock \emph{arXiv preprint arXiv:2403.11481}.

\bibitem[{Fu et~al.(2024)Fu, Dai, Luo, Li, Ren, Zhang, Wang, Zhou, Shen, Zhang et~al.}]{fu2024video}
Chaoyou Fu, Yuhan Dai, Yongdong Luo, Lei Li, Shuhuai Ren, Renrui Zhang, Zihan Wang, Chenyu Zhou, Yunhang Shen, Mengdan Zhang, et~al. 2024.
\newblock Video-mme: The first-ever comprehensive evaluation benchmark of multi-modal llms in video analysis.
\newblock \emph{arXiv preprint arXiv:2405.21075}.

\bibitem[{Fu et~al.(2023)Fu, Li, Gan, Lin, Wang, Wang, and Liu}]{fu2023violet}
Tsu-Jui Fu, Linjie Li, Zhe Gan, Kevin Lin, William~Yang Wang, Lijuan Wang, and Zicheng Liu. 2023.
\newblock An empirical study of end-to-end video-language transformers with masked visual modeling.
\newblock In \emph{Proceedings of the IEEE/CVF Conference on Computer Vision and Pattern Recognition}, pages 22898--22909.

\bibitem[{He et~al.(2016{\natexlab{a}})He, Zhang, Ren, and Sun}]{he2016resnet}
Kaiming He, Xiangyu Zhang, Shaoqing Ren, and Jian Sun. 2016{\natexlab{a}}.
\newblock Deep residual learning for image recognition.
\newblock In \emph{Proceedings of the IEEE conference on computer vision and pattern recognition}, pages 770--778.

\bibitem[{He et~al.(2016{\natexlab{b}})He, Zhang, Ren, and Sun}]{he2016deep}
Kaiming He, Xiangyu Zhang, Shaoqing Ren, and Jian Sun. 2016{\natexlab{b}}.
\newblock Deep residual learning for image recognition.
\newblock In \emph{Proceedings of the IEEE conference on computer vision and pattern recognition}, pages 770--778.

\bibitem[{Hongeng et~al.(2004)Hongeng, Nevatia, and Bremond}]{hongeng2004video}
Somboon Hongeng, Ram Nevatia, and Francois Bremond. 2004.
\newblock Video-based event recognition: activity representation and probabilistic recognition methods.
\newblock \emph{Computer Vision and Image Understanding}, 96(2):129--162.

\bibitem[{Hosseini et~al.(2022)Hosseini, Broniatowski, and Diab}]{lmappce1_hosseini-etal-2022-knowledge}
Pedram Hosseini, David~A. Broniatowski, and Mona Diab. 2022.
\newblock \href {https://doi.org/10.18653/v1/2022.csrr-1.6} {Knowledge-augmented language models for cause-effect relation classification}.
\newblock In \emph{Proceedings of the First Workshop on Commonsense Representation and Reasoning (CSRR 2022)}, pages 43--48, Dublin, Ireland. Association for Computational Linguistics.

\bibitem[{Ivanov and Bobick(2000)}]{ivanov2000recognition}
Yuri~A. Ivanov and Aaron~F. Bobick. 2000.
\newblock Recognition of visual activities and interactions by stochastic parsing.
\newblock \emph{IEEE Transactions on Pattern Analysis and Machine Intelligence}, 22(8):852--872.

\bibitem[{Kahatapitiya et~al.(2024)Kahatapitiya, Ranasinghe, Park, and Ryoo}]{Kahatapitiya2024langrepo}
Kumara Kahatapitiya, Kanchana Ranasinghe, Jongwoo Park, and Michael~S Ryoo. 2024.
\newblock Language repository for long video understanding.

\bibitem[{Kim et~al.(2024)Kim, Choi, Lee, and Rhee}]{kim2024igvlm}
Wonkyun Kim, Changin Choi, Wonseok Lee, and Wonjong Rhee. 2024.
\newblock An image grid can be worth a video: Zero-shot video question answering using a vlm.
\newblock \emph{arXiv preprint arXiv:2403.18406}.

\bibitem[{Lei et~al.(2018)Lei, Yu, Bansal, and Berg}]{dataset_lei2018tvqa}
Jie Lei, Licheng Yu, Mohit Bansal, and Tamara Berg. 2018.
\newblock {TVQA}: Localized, compositional video question answering.
\newblock In \emph{Proceedings of the 2018 Conference on Empirical Methods in Natural Language Processing (EMNLP)}.

\bibitem[{Lei et~al.(2020)Lei, Yu, Berg, and Bansal}]{lei-etal-2020-tvqaplus}
Jie Lei, Licheng Yu, Tamara Berg, and Mohit Bansal. 2020.
\newblock \href {https://doi.org/10.18653/v1/2020.acl-main.730} {{TVQA}+: Spatio-temporal grounding for video question answering}.
\newblock In \emph{Proceedings of the 58th Annual Meeting of the Association for Computational Linguistics}, pages 8211--8225, Online. Association for Computational Linguistics.

\bibitem[{Li et~al.(2024)Li, Zhang, Guo, Zhang, Li, Zhang, Zhang, Zhang, Li, Liu, and Li}]{li2024easyvisualtask-llavaonevision}
Bo~Li, Yuanhan Zhang, Dong Guo, Renrui Zhang, Feng Li, Hao Zhang, Kaichen Zhang, Peiyuan Zhang, Yanwei Li, Ziwei Liu, and Chunyuan Li. 2024.
\newblock \href {https://arxiv.org/abs/2408.03326} {Llava-onevision: Easy visual task transfer}.
\newblock \emph{Preprint}, arXiv:2408.03326.

\bibitem[{Li et~al.(2022)Li, Niu, and Zhang}]{dataset_li2022causalvidqa}
Jiangtong Li, Li~Niu, and Liqing Zhang. 2022.
\newblock From representation to reasoning: Towards both evidence and commonsense reasoning for video question-answering.
\newblock In \emph{Proceedings of the IEEE/CVF Conference on Computer Vision and Pattern Recognition (CVPR)}.

\bibitem[{Li et~al.(2023)Li, Wei, Han, and Fan}]{li2023intentqa_cavir}
Jiapeng Li, Ping Wei, Wenjuan Han, and Lifeng Fan. 2023.
\newblock Intentqa: Context-aware video intent reasoning.
\newblock In \emph{Proceedings of the IEEE/CVF International Conference on Computer Vision}, pages 11963--11974.

\bibitem[{Liu et~al.(2023)Liu, Li, Wu, and Lee}]{liu2023visual}
Haotian Liu, Chunyuan Li, Qingyang Wu, and Yong~Jae Lee. 2023.
\newblock Visual instruction tuning.
\newblock \emph{arXiv preprint arXiv:2304.08485}.

\bibitem[{Maaz et~al.(2023)Maaz, Rasheed, Khan, and Khan}]{Maaz2023VideoChatGPT}
Muhammad Maaz, Hanoona~Abdul Rasheed, Salman~H. Khan, and Fahad~Shahbaz Khan. 2023.
\newblock Video-chatgpt: Towards detailed video understanding via large vision and language models.
\newblock In \emph{Annual Meeting of the Association for Computational Linguistics}.

\bibitem[{Mangalam et~al.(2023)Mangalam, Akshulakov, and Malik}]{Mangalam2023EgoSchemaAD}
Karttikeya Mangalam, Raiymbek Akshulakov, and Jitendra Malik. 2023.
\newblock \href {https://api.semanticscholar.org/CorpusID:261031047} {Egoschema: A diagnostic benchmark for very long-form video language understanding}.
\newblock \emph{ArXiv}, abs/2308.09126.

\bibitem[{Min et~al.(2024)Min, Buch, Nagrani, Cho, and Schmid}]{min2024morevqa}
Juhong Min, Shyamal Buch, Arsha Nagrani, Minsu Cho, and Cordelia Schmid. 2024.
\newblock Morevqa: Exploring modular reasoning models for video question answering.
\newblock In \emph{Proceedings of the IEEE/CVF Conference on Computer Vision and Pattern Recognition}, pages 13235--13245.

\bibitem[{Momeni et~al.(2023)Momeni, Caron, Nagrani, Zisserman, and Schmid}]{momeni2023vfc}
Liliane Momeni, Mathilde Caron, Arsha Nagrani, Andrew Zisserman, and Cordelia Schmid. 2023.
\newblock Verbs in action: Improving verb understanding in video-language models.
\newblock In \emph{Proceedings of the IEEE/CVF International Conference on Computer Vision}, pages 15579--15591.

\bibitem[{Papalampidi et~al.(2023)Papalampidi, Koppula, Pathak, Chiu, Heyward, Patraucean, Shen, Miech, Zisserman, and Nematzdeh}]{papalampidi2023simple}
Pinelopi Papalampidi, Skanda Koppula, Shreya Pathak, Justin Chiu, Joe Heyward, Viorica Patraucean, Jiajun Shen, Antoine Miech, Andrew Zisserman, and Aida Nematzdeh. 2023.
\newblock A simple recipe for contrastively pre-training video-first encoders beyond 16 frames.
\newblock \emph{arXiv preprint arXiv:2312.07395}.

\bibitem[{Ranasinghe et~al.(2024)Ranasinghe, Li, Kahatapitiya, and Ryoo}]{rana2024mvu}
Kanchana Ranasinghe, Xiang Li, Kumara Kahatapitiya, and Michael Ryoo. 2024.
\newblock Understanding long videos in one multimodal language model pass.

\bibitem[{Ranasinghe et~al.(2023)Ranasinghe, McKinzie, Ravi, Yang, Toshev, and Shlens}]{Ranasinghe2022PerceptualGI}
Kanchana Ranasinghe, Brandon McKinzie, Sachin Ravi, Yinfei Yang, Alexander Toshev, and Jonathon Shlens. 2023.
\newblock Perceptual grouping in contrastive vision-language models.
\newblock In \emph{ICCV}.

\bibitem[{Ryoo and Aggarwal(2006)}]{Ryoo2006RecognitionOC}
Michael~S. Ryoo and Jake~K. Aggarwal. 2006.
\newblock \href {https://api.semanticscholar.org/CorpusID:14039104} {Recognition of composite human activities through context-free grammar based representation}.
\newblock \emph{2006 IEEE Computer Society Conference on Computer Vision and Pattern Recognition (CVPR'06)}, 2:1709--1718.

\bibitem[{Shang et~al.(2024)Shang, You, Subramanian, Darrell, and Herzig}]{shang2024traveler}
Chuyi Shang, Amos You, Sanjay Subramanian, Trevor Darrell, and Roei Herzig. 2024.
\newblock Traveler: A modular multi-lmm agent framework for video question-answering.
\newblock \emph{arXiv preprint arXiv:2404.01476}.

\bibitem[{Shi et~al.(2004)Shi, Huang, Minnen, Bobick, and Essa}]{shi2004propagation}
Yifan Shi, Yan Huang, David Minnen, Aaron Bobick, and Irfan Essa. 2004.
\newblock Propagation networks for recognition of partially ordered sequential action.
\newblock In \emph{CVPR}.

\bibitem[{Tapaswi et~al.(2016)Tapaswi, Zhu, Stiefelhagen, Torralba, Urtasun, and Fidler}]{dataset_tapaswi2016movieqa}
Makarand Tapaswi, Yukun Zhu, Rainer Stiefelhagen, Antonio Torralba, Raquel Urtasun, and Sanja Fidler. 2016.
\newblock {MovieQA}: Understanding stories in movies through question-answering.
\newblock In \emph{Proceedings of the IEEE/CVF Conference on Computer Vision and Pattern Recognition (CVPR)}.

\bibitem[{Wang et~al.(2024{\natexlab{a}})Wang, Yuan, and Zhang}]{wang2024tarsier}
Jiawei Wang, Liping Yuan, and Yuchen Zhang. 2024{\natexlab{a}}.
\newblock Tarsier: Recipes for training and evaluating large video description models.
\newblock \emph{arXiv preprint arXiv:2407.00634}.

\bibitem[{Wang et~al.(2024{\natexlab{b}})Wang, Bai, Tan, Wang, Fan, Bai, Chen, Liu, Wang, Ge, Fan, Dang, Du, Ren, Men, Liu, Zhou, Zhou, and Lin}]{Wang2024Qwen2VLEV}
Peng Wang, Shuai Bai, Sinan Tan, Shijie Wang, Zhihao Fan, Jinze Bai, Ke-Yang Chen, Xuejing Liu, Jialin Wang, Wenbin Ge, Yang Fan, Kai Dang, Mengfei Du, Xuancheng Ren, Rui Men, Dayiheng Liu, Chang Zhou, Jingren Zhou, and Junyang Lin. 2024{\natexlab{b}}.
\newblock \href {https://api.semanticscholar.org/CorpusID:272704132} {Qwen2-vl: Enhancing vision-language model's perception of the world at any resolution}.
\newblock \emph{ArXiv}, abs/2409.12191.

\bibitem[{Wang et~al.(2024{\natexlab{c}})Wang, Bai, Tan, Wang, Fan, Bai, Chen, Liu, Wang, Ge, Fan, Dang, Du, Ren, Men, Liu, Zhou, Zhou, and Lin}]{Qwen2-VL}
Peng Wang, Shuai Bai, Sinan Tan, Shijie Wang, Zhihao Fan, Jinze Bai, Keqin Chen, Xuejing Liu, Jialin Wang, Wenbin Ge, Yang Fan, Kai Dang, Mengfei Du, Xuancheng Ren, Rui Men, Dayiheng Liu, Chang Zhou, Jingren Zhou, and Junyang Lin. 2024{\natexlab{c}}.
\newblock Qwen2-vl: Enhancing vision-language model's perception of the world at any resolution.
\newblock \emph{arXiv preprint arXiv:2409.12191}.

\bibitem[{Wang et~al.(2023)Wang, Zhao, Do, Agarwal, Lee, and Sun}]{wang2023vamos}
Shijie Wang, Qi~Zhao, Minh~Quan Do, Nakul Agarwal, Kwonjoon Lee, and Chen Sun. 2023.
\newblock Vamos: Versatile action models for video understanding.
\newblock \emph{arXiv preprint arXiv:2311.13627}.

\bibitem[{Wang et~al.(2024{\natexlab{d}})Wang, Zhang, Zohar, and Yeung-Levy}]{Wang2024VideoAgentLV}
Xiaohan Wang, Yuhui Zhang, Orr Zohar, and Serena Yeung-Levy. 2024{\natexlab{d}}.
\newblock Videoagent: Long-form video understanding with large language model as agent.

\bibitem[{Wang et~al.(2024{\natexlab{e}})Wang, Li, Li, Yu, He, Chen, Pei, Zheng, Xu, Wang et~al.}]{wang2024internvideo2}
Yi~Wang, Kunchang Li, Xinhao Li, Jiashuo Yu, Yinan He, Guo Chen, Baoqi Pei, Rongkun Zheng, Jilan Xu, Zun Wang, et~al. 2024{\natexlab{e}}.
\newblock Internvideo2: Scaling video foundation models for multimodal video understanding.
\newblock \emph{arXiv preprint arXiv:2403.15377}.

\bibitem[{Wang et~al.(2024{\natexlab{f}})Wang, Yang, and Ren}]{wang2024lifelongmemory}
Ying Wang, Yanlai Yang, and Mengye Ren. 2024{\natexlab{f}}.
\newblock \href {https://arxiv.org/abs/2312.05269} {Lifelongmemory: Leveraging llms for answering queries in long-form egocentric videos}.
\newblock \emph{Preprint}, arXiv:2312.05269.

\bibitem[{Wang et~al.(2024{\natexlab{g}})Wang, Yu, Stengel-Eskin, Yoon, Cheng, Bertasius, and Bansal}]{wang2024videotree}
Ziyang Wang, Shoubin Yu, Elias Stengel-Eskin, Jaehong Yoon, Feng Cheng, Gedas Bertasius, and Mohit Bansal. 2024{\natexlab{g}}.
\newblock Videotree: Adaptive tree-based video representation for llm reasoning on long videos.
\newblock \emph{arXiv preprint arXiv:2405.19209}.

\bibitem[{Xiao et~al.(2021)Xiao, Shang, Yao, and Chua}]{dataset_xiao2021nextqa}
Junbin Xiao, Xindi Shang, Angela Yao, and Tat-Seng Chua. 2021.
\newblock {NExT-QA}: Next phase of question-answering to explaining temporal actions.
\newblock In \emph{Proceedings of the IEEE/CVF Conference on Computer Vision and Pattern Recognition (CVPR)}.

\bibitem[{Xiao et~al.(2022{\natexlab{a}})Xiao, Yao, Liu, Li, Ji, and Chua}]{model_xiao2021hgqa}
Junbin Xiao, Angela Yao, Zhiyuan Liu, Yicong Li, Wei Ji, and Tat-Seng Chua. 2022{\natexlab{a}}.
\newblock Video as conditional graph hierarchy for multi-granular question answering.
\newblock In \emph{Proceedings of the 36th AAAI Conference on Artificial Intelligence (AAAI)}, pages 2804--2812.

\bibitem[{Xiao et~al.(2022{\natexlab{b}})Xiao, Zhou, Chua, and Yan}]{model_xiao2022vgt}
Junbin Xiao, Pan Zhou, Tat-Seng Chua, and Shuicheng Yan. 2022{\natexlab{b}}.
\newblock Video graph transformer for video question answering.
\newblock In \emph{European Conference on Computer Vision}, pages 39--58. Springer.

\bibitem[{Xu et~al.(2017)Xu, Zhao, Xiao, Wu, Zhang, He, and Zhuang}]{xu2017msvdqavideo}
Dejing Xu, Zhou Zhao, Jun Xiao, Fei Wu, Hanwang Zhang, Xiangnan He, and Yueting Zhuang. 2017.
\newblock Video question answering via gradually refined attention over appearance and motion.
\newblock In \emph{ACM Multimedia}.

\bibitem[{Yang et~al.(2022)Yang, Miech, Sivic, Laptev, and Schmid}]{yang2022frozenblim}
Antoine Yang, Antoine Miech, Josef Sivic, Ivan Laptev, and Cordelia Schmid. 2022.
\newblock Zero-shot video question answering via frozen bidirectional language models.
\newblock \emph{Advances in Neural Information Processing Systems}, 35:124--141.

\bibitem[{Ye et~al.(2023)Ye, Xu, Xu, Ye, Yan, Zhou, Wang, Hu, Shi, Shi et~al.}]{ye2023mplug}
Qinghao Ye, Haiyang Xu, Guohai Xu, Jiabo Ye, Ming Yan, Yiyang Zhou, Junyang Wang, Anwen Hu, Pengcheng Shi, Yaya Shi, et~al. 2023.
\newblock mplug-owl: Modularization empowers large language models with multimodality.
\newblock \emph{arXiv preprint arXiv:2304.14178}.

\bibitem[{Yu et~al.(2023)Yu, Cho, Yadav, and Bansal}]{Yu2023SelfChainedIM}
Shoubin Yu, Jaemin Cho, Prateek Yadav, and Mohit Bansal. 2023.
\newblock \href {https://api.semanticscholar.org/CorpusID:258615748} {Self-chained image-language model for video localization and question answering}.
\newblock \emph{ArXiv}, abs/2305.06988.

\bibitem[{Yu et~al.(2024{\natexlab{a}})Yu, Cho, Yadav, and Bansal}]{yu2024sevila}
Shoubin Yu, Jaemin Cho, Prateek Yadav, and Mohit Bansal. 2024{\natexlab{a}}.
\newblock Self-chained image-language model for video localization and question answering.
\newblock \emph{Advances in Neural Information Processing Systems}, 36.

\bibitem[{Yu et~al.(2024{\natexlab{b}})Yu, Jin, Wang, Chen, Jin, Zuo, Xu, Sun, Zhang, Wu et~al.}]{yu2024framevoyager}
Sicheng Yu, Chengkai Jin, Huanyu Wang, Zhenghao Chen, Sheng Jin, Zhongrong Zuo, Xiaolei Xu, Zhenbang Sun, Bingni Zhang, Jiawei Wu, et~al. 2024{\natexlab{b}}.
\newblock Frame-voyager: Learning to query frames for video large language models.
\newblock \emph{arXiv preprint arXiv:2410.03226}.

\bibitem[{Yu et~al.(2019{\natexlab{a}})Yu, Xu, Yu, Yu, Zhao, Zhuang, and Tao}]{Yu2019ActivityNetQAAD}
Zhou Yu, D.~Xu, Jun Yu, Ting Yu, Zhou Zhao, Yueting Zhuang, and Dacheng Tao. 2019{\natexlab{a}}.
\newblock \href {https://api.semanticscholar.org/CorpusID:69645185} {Activitynet-qa: A dataset for understanding complex web videos via question answering}.
\newblock \emph{ArXiv}, abs/1906.02467.

\bibitem[{Yu et~al.(2019{\natexlab{b}})Yu, Xu, Yu, Yu, Zhao, Zhuang, and Tao}]{dataset_anetqa}
Zhou Yu, Dejing Xu, Jun Yu, Ting Yu, Zhou Zhao, Yueting Zhuang, and Dacheng Tao. 2019{\natexlab{b}}.
\newblock {ActivityNet-QA}: A dataset for understanding complex web videos via question answering.
\newblock In \emph{Proceedings of the AAAI Conference on Artificial Intelligence (AAAI)}.

\bibitem[{Zeng et~al.(2017)Zeng, Chen, Chuang, Liao, Niebles, and Sun}]{videoqaaaai2017}
Kuo-Hao Zeng, Tseng-Hung Chen, Ching-Yao Chuang, Yuan-Hong Liao, Juan~Carlos Niebles, and Min Sun. 2017.
\newblock \href {https://api.semanticscholar.org/CorpusID:7224807} {Leveraging video descriptions to learn video question answering}.
\newblock In \emph{AAAI Conference on Artificial Intelligence}.

\bibitem[{Zeng et~al.(2024)Zeng, Li, Wang, Li, Jiang, Yan, Li, Shi, Yue, Wang et~al.}]{zeng2024timesuite}
Xiangyu Zeng, Kunchang Li, Chenting Wang, Xinhao Li, Tianxiang Jiang, Ziang Yan, Songze Li, Yansong Shi, Zhengrong Yue, Yi~Wang, et~al. 2024.
\newblock Timesuite: Improving mllms for long video understanding via grounded tuning.
\newblock \emph{arXiv preprint arXiv:2410.19702}.

\bibitem[{Zhang et~al.(2023)Zhang, Lu, Islam, Wang, Yu, Bansal, and Bertasius}]{zhang2023llovi}
Ce~Zhang, Taixi Lu, Md~Mohaiminul Islam, Ziyang Wang, Shoubin Yu, Mohit Bansal, and Gedas Bertasius. 2023.
\newblock A simple llm framework for long-range video question-answering.
\newblock \emph{arXiv preprint arXiv:2312.17235}.

\bibitem[{Zhang et~al.(2024{\natexlab{a}})Zhang, Wu, Li, Li, Ma, Liu, and Li}]{zhang2024videoinstructiontuningsynthetic-llavanext}
Yuanhan Zhang, Jinming Wu, Wei Li, Bo~Li, Zejun Ma, Ziwei Liu, and Chunyuan Li. 2024{\natexlab{a}}.
\newblock \href {https://arxiv.org/abs/2410.02713} {Video instruction tuning with synthetic data}.
\newblock \emph{Preprint}, arXiv:2410.02713.

\bibitem[{Zhang et~al.(2024{\natexlab{b}})Zhang, Wu, Li, Li, Ma, Liu, and Li}]{zhang2024llava-video}
Yuanhan Zhang, Jinming Wu, Wei Li, Bo~Li, Zejun Ma, Ziwei Liu, and Chunyuan Li. 2024{\natexlab{b}}.
\newblock \href {https://arxiv.org/abs/2410.02713} {Video instruction tuning with synthetic data}.
\newblock \emph{Preprint}, arXiv:2410.02713.

\bibitem[{Zhao et~al.(2017)Zhao, Ma, and You}]{Zhao_2017_ICCV}
Zhichen Zhao, Huimin Ma, and Shaodi You. 2017.
\newblock Single image action recognition using semantic body part actions.
\newblock In \emph{The IEEE International Conference on Computer Vision (ICCV)}.

\bibitem[{Zhou et~al.(2024)Zhou, Shu, Zhao, Wu, Xiao, Yang, Xiong, Zhang, Huang, and Liu}]{MLVU}
Junjie Zhou, Yan Shu, Bo~Zhao, Boya Wu, Shitao Xiao, Xi~Yang, Yongping Xiong, Bo~Zhang, Tiejun Huang, and Zheng Liu. 2024.
\newblock Mlvu: A comprehensive benchmark for multi-task long video understanding.
\newblock \emph{arXiv preprint arXiv:2406.04264}.

\end{thebibliography}

\newpage
\clearpage
\appendix
\section*{Appendix}

\renewcommand{\thetable}{A.\arabic{table}}
\renewcommand{\thefigure}{A.\arabic{figure}}
\renewcommand{\thesection}{A.\arabic{section}}
\renewcommand{\thesubsection}{A.\arabic{section}.\arabic{subsection}}

\begin{figure*}
\centering
\includegraphics[width=0.95\linewidth]{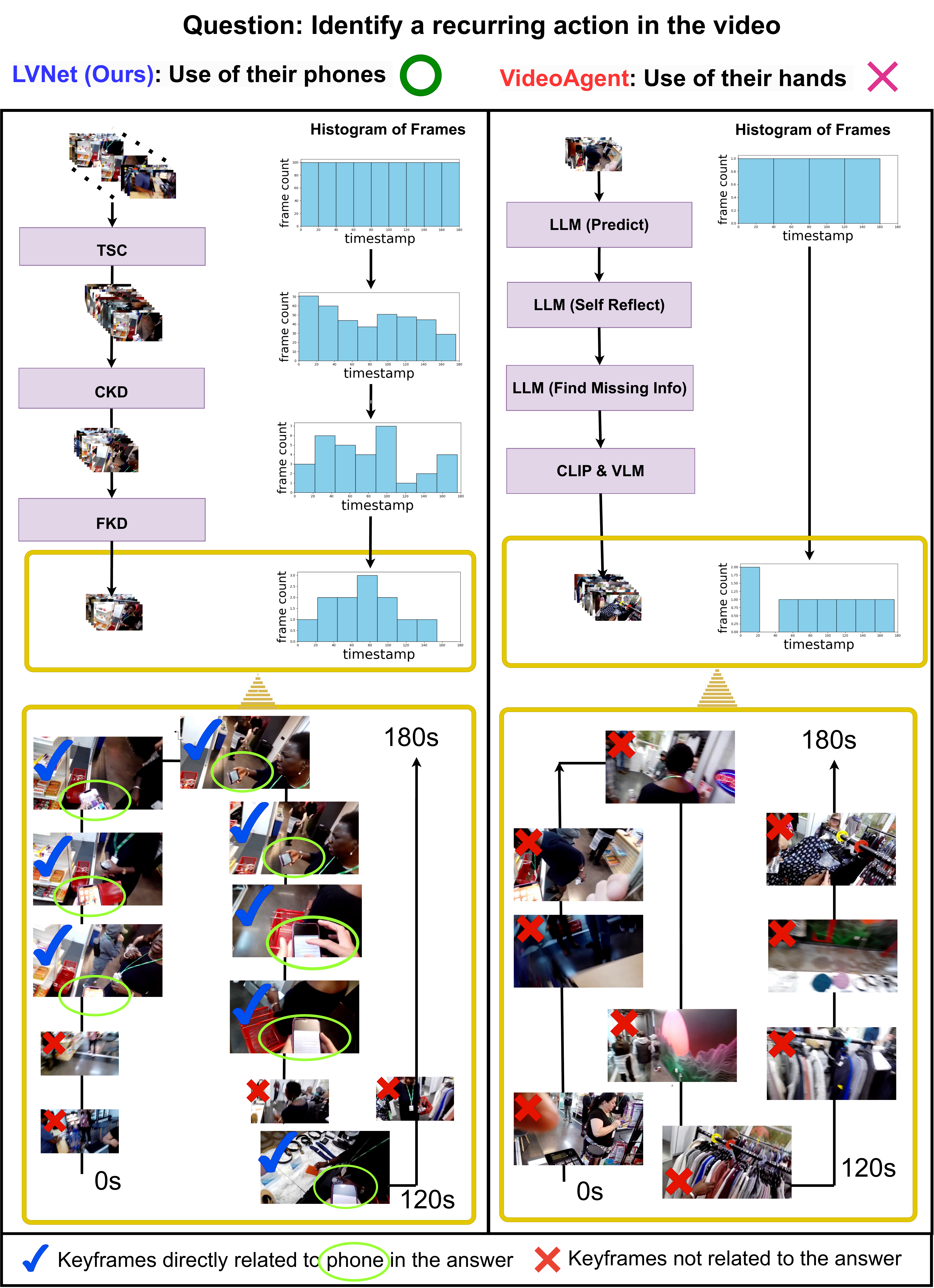}
\caption{\textbf{Comparison of Keyframe Selection}: Comparison of LVNet and VideoAgent in keyframe selection for video question answering. LVNet refines frames through a multi-stage process (TSC, CKD, FKD) to form a non-uniform keyframe distribution, capturing relevant moments tied to the query. In contrast, VideoAgent relies on uniform sampling and LLM-based frame selection, which fails to focus on crucial keyframes, leading to incorrect predictions. The final keyframe distributions illustrate LVNet's ability to retrieve meaningful frames directly related to the answer, while VideoAgent selects irrelevant frames.
}\label{fig:kfscomparison4}
\end{figure*}

\section{Additional Ablations}
In this section, we present additional experiments conducted to inform the \modelname's design. We have tested different LLMs and experimented with various scales of the visual feature map.
\label{app:additionalablation}

\begin{table}[ht]\centering
    \label{res:supp_ablate}\vspace{3mm}
    \subfloat[\textbf{Uniform vs LVNet}. \label{tab:appendix_uniform}
    ]{
        \resizebox{0.48\columnwidth}{!}{
        \begin{tabular}{lcc}
        \toprule
        Model & Frames & Acc. (\%) \\
        \midrule
        GPT\mbox{-}4o & 12 & 62.6 \\
        \midrule
        LVNet  & 8  & 64.4 \\
        \rowcolor{row}
        LVNet  & 12 & 68.2 \\
        \bottomrule
        \end{tabular}}}\hspace{3mm}
    \subfloat[\textbf{Effect of Patch Size in CKD}: A larger patch size in Keyword Matching performs better. \label{tab:ablation:patchsizeinckd}
    ]{
        \resizebox{0.42\columnwidth}{!}{
        \begin{tabular}{lc}
        \toprule
          \multicolumn{1}{l}{Patch Size$\;\;\;$} & Acc. (\%) \\
        \midrule
         1x1 & 63.6  \\
         7x7 & 66.2 \\
         \rowcolor{row}
         14x14 & 68.2 \\
        \bottomrule
        \end{tabular}}}
    \vspace{-2mm}
    \caption{\textbf{Additional comparisons} on EgoSchema subset. Left: uniform 12‑frame sampling (GPT‑4o) vs LVNet at 8 and 12 frames. Right: effect of patch size in CKD.}
\end{table}

\paragraph{Uniform Sampling Comparison:}
GPT\mbox{-}4o with uniform 12‑frame sampling on the EgoSchema subset scores 62.6\%. LVNet outperforms GPT\mbox{-}4o, reaching 64.4\% with just 8 frames and 68.2\% with 12 frames (Table~\ref{tab:appendix_uniform}). To keep the evaluation both methodical and cost‑effective, we first ran GPT\mbox{-}4o and LVNet on the EgoSchema subset as validation to find the optimal settings, then transferred those settings to the full EgoSchema test set and the remaining benchmarks. This procedure avoids the $>\$3{,}000$ API expense of running GPT\mbox{-}4o uniformly across all four datasets while still demonstrating LVNet’s stronger generalization. For LLM choices, see Sec.~\ref{subsec:efficiency_scalability}.

\vspace{0.2em}
\paragraph{Effect of Patch Size on Keyword Matching in CKD:} \tref{tab:ablation:patchsizeinckd} shows the effect of the scales of the patch sizes in the CKD. Since keywords can represent activities spanning the entire image or confined to a small region, we adjust the resolution of the visual feature map output from the spatially aware contrastive image pre-training (CLIP) network \citep{Ranasinghe2022PerceptualGI} to match keywords. Our findings show that higher resolutions lead to better accuracy. In \modelname, we use a 14$\times$14 feature map and determine the confidence level of the keyword by selecting the maximum value between the 14$\times$14 patches and the keyword's text embedding.

\section{Extended results on NExT-QA and IntentQA}
\label{app:extended_res}

We present extended zero-shot evaluation results on NExT-QA in \Cref{tab:supp_nextqa}, comparing \modelname~with prior zero-shot models across different task categories: causal, temporal, and descriptive reasoning. Models are ordered based on the number of captions processed per video, highlighting the trade-offs between caption efficiency and performance.

\modelname~achieves state-of-the-art performance with an overall accuracy of 72.9\%, outperforming most models while using only 12 captions per video. Notably, it attains 75.0\% on causal reasoning, which is the highest among all models evaluated. For temporal reasoning, \modelname~achieves 65.5\%, remaining competitive despite using significantly fewer captions than models like VideoTree (56 captions) and LangRepo (90 captions). In descriptive reasoning, \modelname~reaches 81.5\%, matching VideoTree while processing significantly fewer captions.

Compared to VideoAgent, the closest competing model in terms of caption efficiency (8.4 captions), \modelname~demonstrates a substantial performance gain across all categories, with a +2.8\% improvement in overall accuracy. While models like VideoTree and TraveLER show strong performance, they process significantly more captions (56 and 65, respectively), indicating that \modelname~achieves a superior balance between efficiency and accuracy.

We present extended zero-shot evaluation results on IntentQA in \Cref{tab:supp_intentqa}, comparing \modelname~with prior zero-shot models across different reasoning categories: \textit{Why?}, \textit{How?}, and \textit{B.A.} (Before/After). Models are ordered based on the number of captions processed per video, highlighting the balance between caption efficiency and performance.

\modelname~achieves an overall accuracy of 71.7\%, outperforming all models while using only 12 captions per video. It achieves 75.0\% on the \textit{Why?} category, 74.4\% on the \textit{How?} category, and 62.1\% on the \textit{B.A.} category. Compared to VideoTree, which processes 56 captions and achieves an overall accuracy of 66.9\%, \modelname~outperforms it by +4.8\% while using significantly fewer captions. Similarly, LangRepo and LLoVi, which process 90 captions, achieve overall scores of 59.1\% and 64.0\%, respectively, further demonstrating \modelname's caption efficiency.

To ensure fairness, models that utilize video-caption pretraining or process substantially more captions than \modelname~are \textcolor{down}{de-emphasized in grey} or \textcolor{down2}{downplayed in light green} in \Cref{tab:supp_nextqa} and \Cref{tab:supp_intentqa}.

\begin{table*}[ht]
\centering
\resizebox{0.75\textwidth}{!}{
\begin{tabular}{lccccc}
\toprule
\multirow{1}{*}{Model} & Cap. & Cau. (\%) & Tem. (\%) & Des. (\%) & All (\%) \\ 
\midrule
IG-VLM \citep{kim2024igvlm} & - & 69.8 & 63.6 & 74.7 & 68.6 \\
\textcolor{down}{Tarsier \citep{wang2024tarsier}} & \textcolor{down}{-} & \textcolor{down}{-} & \textcolor{down}{-} & \textcolor{down}{-} & \textcolor{down}{79.2} \\
VideoAgent \citep{Wang2024VideoAgentLV} & 8.2 & 72.7 & 64.5 & 81.1 & 71.3 \\
MVU \citep{rana2024mvu} & 16 & 55.4 & 48.1 & 64.1 & 55.2 \\
MoReVQA \citep{min2024morevqa} & 16 & 70.2 & 64.6 & - & 69.2 \\
VFC \citep{momeni2023vfc} & 32 & 45.4 & 51.6 & 64.1 & 51.5 \\
\textcolor{down}{SeViLA$^\dagger$ \citep{yu2024sevila}} & \textcolor{down}{32} & \textcolor{down}{61.3} & \textcolor{down}{61.5} & \textcolor{down}{75.6} & \textcolor{down}{63.6} \\
\textcolor{down2}{VideoTree \citep{wang2024videotree}} & \textcolor{down2}{(56)} & \textcolor{down2}{75.2} & \textcolor{down2}{67.0} & \textcolor{down2}{81.3} & \textcolor{down2}{73.5} \\
\textcolor{down2}{ProViQ \citep{choudhury2023proviq}} & \textcolor{down2}{60} & \textcolor{down2}{-} & \textcolor{down2}{-} & \textcolor{down2}{-} & \textcolor{down2}{64.6}\\
\textcolor{down2}{TraveLER} \citep{shang2024traveler} & \textcolor{down2}{(65)} & \textcolor{down2}{70.0} & \textcolor{down2}{60.5} & \textcolor{down2}{78.2} & \textcolor{down2}{68.2} \\
\textcolor{down2}{LangRepo \citep{Kahatapitiya2024langrepo}} & \textcolor{down2}{90} & \textcolor{down2}{64.4} & \textcolor{down2}{51.4} & \textcolor{down2}{69.1} & \textcolor{down2}{60.9} \\
\textcolor{down2}{LLoVi \citep{zhang2023llovi}} & \textcolor{down2}{90} & \textcolor{down2}{69.5} & \textcolor{down2}{61.0} & \textcolor{down2}{75.6} & \textcolor{down2}{67.7} \\
\midrule
\rowcolor{row}\modelname~(ours) & 12 & 75.0 & 65.5 & 81.5 & 72.9 \\
\bottomrule
\end{tabular}
}
\caption{
\textbf{Extended results on NExT-QA \citep{dataset_xiao2021nextqa}.} 
We compare \modelname~against prior zero-shot models across different reasoning categories: causal, temporal, and descriptive. \modelname~achieves an overall accuracy of 72.9\% while using only 12 captions per video, demonstrating strong performance across all reasoning types. Notably, it outperforms all models in causal reasoning (75.0\%) and matches the best performance in descriptive reasoning (81.5\%), despite processing significantly fewer captions than models like VideoTree (56 captions) and TraveLER (65 captions). Models that utilize video-caption pretraining or process substantially more captions than \modelname~are \textcolor{down}{de-emphasized in gray} or \textcolor{down2}{downplayed in light green} to ensure fairness in comparison. Numbers in parentheses () indicate the maximum number of frames used.
}
\vspace{-0.5em}
\label{tab:supp_nextqa}
\end{table*}

\begin{table*}[ht]
\centering
\resizebox{0.75\textwidth}{!}{
\begin{tabular}{lccccc}
\toprule
\multirow{1}{*}{Model} & Cap. & Why? (\%) & How? (\%) & B./A. (\%) & All (\%) \\ 
\midrule
IG-VLM \citep{kim2024igvlm} & - & - & - & - & 65.3 \\

\textcolor{down}{SeViLA$^\dagger$ \citep{yu2024sevila}} & \textcolor{down}{32} & \textcolor{down}{-} & \textcolor{down}{-} & \textcolor{down}{-} & \textcolor{down}{60.9} \\

\textcolor{down2}{VideoTree \citep{wang2024videotree}} & \textcolor{down2}{(56)} & \textcolor{down2}{-} & \textcolor{down2}{-} & \textcolor{down2}{-} & \textcolor{down2}{66.9} \\

\textcolor{down2}{LangRepo \citep{Kahatapitiya2024langrepo}} & \textcolor{down2}{90} & \textcolor{down2}{62.8} & \textcolor{down2}{62.4} & \textcolor{down2}{47.8} & \textcolor{down2}{59.1} \\
\textcolor{down2}{LLoVi \citep{zhang2023llovi}} & \textcolor{down2}{90} & \textcolor{down2}{68.4} & \textcolor{down2}{67.4} & \textcolor{down2}{51.1} & \textcolor{down2}{64.0} \\
\midrule
\rowcolor{row}\modelname~(ours) & 12 & 75.0 & 74.4 & 62.1 & 71.7 \\
\bottomrule
\end{tabular}
}
\caption{
\textbf{Extended results on IntentQA \citep{li2023intentqa_cavir}.} 
We compare \modelname~against prior zero-shot models across different reasoning categories: \textit{Why?}, \textit{How?}, and \textit{B.A.} (Belief/Action). \modelname~achieves an overall accuracy of 71.7\%, surpassing all models while using only 12 captions per video. It reaches 75.0\% in the \textit{Why?} category, 74.4\% in the \textit{How?} category, and 62.1\% in the \textit{B.A.} category. Compared to VideoTree, which processes 56 captions and achieves 66.9\% accuracy, \modelname~outperforms it by +4.8\% while using significantly fewer captions. Additionally, \modelname~demonstrates superior reasoning-based performance compared to LangRepo (90 captions, 59.1\%) and LLoVi (90 captions, 64.0\%). Models with video-caption pretraining or utilizing significantly more captions than 12 frames used by \modelname~are \textcolor{down}{de-emphasized in grey} or \textcolor{down2}{downplayed in light green} to ensure fairness with image-level pretraining or highlight caption efficiency. Numbers in parentheses () indicate the maximum number of frames used.
}
\vspace{-0.5em}
\label{tab:supp_intentqa}
\end{table*}

\section{Extended results on VideoMME}
\label{app:extended_videomme}
\begin{table*}[ht]
\centering
\resizebox{1.0\textwidth}{!}{
\begin{tabular}{l c c c c c c c c}
\toprule
\textbf{Method} & \textbf{LLM Active Params/Type} & \textbf{TS} & \textbf{VT Free} & \textbf{\# Frames (n)}$\uparrow$ & \textbf{TS Cap.} $\downarrow$& \textbf{FR} $\uparrow$ & \textbf{Complexity} & \textbf{Accuracy} $\uparrow$ \\
\midrule
Qwen-VL \citep{bai2023versatilevisionlanguagemodel-qwenvl}              & 7B/OS         & no & no  & 4    & N/A  & --  & --    & 37.8 \\
Qwen-VL              & 7B/OS         & no & no  & 1800 & N/A & --  & --     & OOM  \\
LongVILA \citep{chen2024scalinglongcontextvisual-longvila}            & 8B/OS         & no & no  & 128  & N/A   & -- & --   & 38.8 \\
LongVILA             & 8B/OS         & no & no  & 256  & N/A & --  & --    & 39.7 \\
LLaVA-OneVision \citep{li2024easyvisualtask-llavaonevision}     & 7B/OS         & no & no  & 8    & N/A & --  & --   & 43.8 \\
VideoChat-T \citep{zeng2024timesuite}         & 7B/OS         & no & no  & 128  & N/A & 87.5 & \(\mathcal{O}(n)\)   & 41.9 \\
Frame-Voyager \citep{yu2024framevoyager}       & 7B/OS         & no & no  & 128  & N/A & 93.7 & \(\mathcal{O}{(n\textsuperscript{r})}\) & 48.9 \\
LLaVA-NexT-Video \citep{zhang2024videoinstructiontuningsynthetic-llavanext}     & 34B/OS        & no & no  & 32   & N/A & --  & --    & 44.3 \\
Frame-Voyager        & 34B/OS        & no & no  & 128  & N/A & 93.7  &  \(\mathcal{O}{(n\textsuperscript{r})}\) & 51.2 \\
InternVL2 \citep{chen2024far_internvl2} & 34B/OS & no & no & 16 & N/A & -- & -- & 52.6 \\
\rowcolor{row}LVNet (\small{DeepSeek-v3}) & 37B/OS        & yes  & yes & 1800 & 24 & --  & \(\mathcal{O}(n)\)   & 53.1 \\
\midrule
VideoAgent+\small{GPT4o}           & <1.8T/PP   & yes  & yes & --   & 24 & -- & \(\mathcal{O}(m)\)  & 51.3 \\
VideoTree+\small{GPT4o}         & <1.8T/PP   & yes  & yes & 300  & 24 & 92.0 & \(\mathcal{O}{(n\textsuperscript{r})}\) & 52.3 \\
GPT4o direct             & <1.8T/PP   & yes  & yes  & 1800 & 1800 & -- & --     & API Error \\
\rowcolor{row}LVNet (\small{\gpto})      & <1.8T/PP   & yes  & yes & 1800 & 24 &  98.7 & \(\mathcal{O}(n)\)   & 53.9 \\
\midrule
LLaVA-Video \citep{zhang2024llava-video}             & 72B/OS         & no & no  & 64    & N/A  & -- & --     & 61.5 \\
Qwen2-VL \citep{Qwen2-VL}            & 72B/OS         & no & no  & 768    & N/A & -- & --     & 62.2 \\
InternVL2.5  \citep{chen2024intervl25}            & 72B/OS         & no & no  & 64    & N/A  & -- & --     & 62.6 \\
\bottomrule
\end{tabular}
}
\vspace{-0.5em}
\caption{\textbf{Detailed Comparison on VideoMME \citep{fu2024video}} . This table expands on \Cref{tab:vidmme} by including two-stage (TS), video-level training free (VT Free), number of input frames used, two-stage captions fed to LLM (TS Cap.), ratio of frames/captions provided to the LLM relative to input frames (FR), and frame selection complexity. “OS” denotes open-source, "OOM” stands for out-of-memory. \(\mathcal{O}(m)\) denotes the complexity proportional to the number LLM calls to predict keyframe timelines. The bottom 3 results are from benchmark leaderboard; we are unable to replicate these on our compute resources.
Our LVNet variants require no video-level training yet achieve competitive results on very long videos.
}
\label{tab:extendedvideomme}
\end{table*}

\Cref{tab:extendedvideomme} provides an extended comparison of both \emph{single-stage} and \emph{two-stage} approaches on the VideoMME long-video benchmark. In contrast to single-stage methods, which typically require costly video-level training on large datasets which has the risk of containing evaluation datasets. LVNet is video-level training free and thus retains broader generality. Critically, single-stage pipelines often become infeasible for long videos because they must process entire sequences—potentially exceeding 1,000 frames—through a heavy vision-language model (VLM) or LLM.

Keyframe-selection approaches mitigate this challenge by filtering a minimal set of relevant frames before any large-model processing. As shown in \Cref{tab:extendedvideomme}, LVNet can handle up to 1,800 frames by reducing them to just a few dozen keyframes, thereby remaining both memory- and compute-efficient while still achieving strong accuracy. For Qwen-VL and GPT4o, we demonstrate an ablation with 1,800 frames to highlight how easily single-stage methods can run out of memory (OOM) under 4 NVIDIA RTX A5000 GPUs or encounter GPT API errors.

Moreover, our approach achieves strong performance with both open-source and proprietary LLM backbones, all without requiring any video-level training. LVNet+DeepSeek-V3 outperforms these single-stage keyframe selection models (VideoChat-T, Frame-Voyager) with equal or less frame selection complexity and outperforms single-stage VideoLLMs with a similarly sized open-source LLM. In a direct two-stage head-to-head setting that uses \gpto{} for both frame captioning and LLM reasoning and enforces the same 24-caption budget, LVNet further surpasses VideoAgent and VideoTree by +2.6\% and +1.6\%, respectively. Taken together, these results show LVNet's effectiveness over both single-stage and two-stage video methods. The accuracy comes from each model’s recommended configuration and matched budgets for fair comparison.

\section{Qualitative Analysis of Hierarchical Keyframe Selector}
\label{app:qualitative}
\begin{figure*}[th]
\centering
\includegraphics[width=1\linewidth]{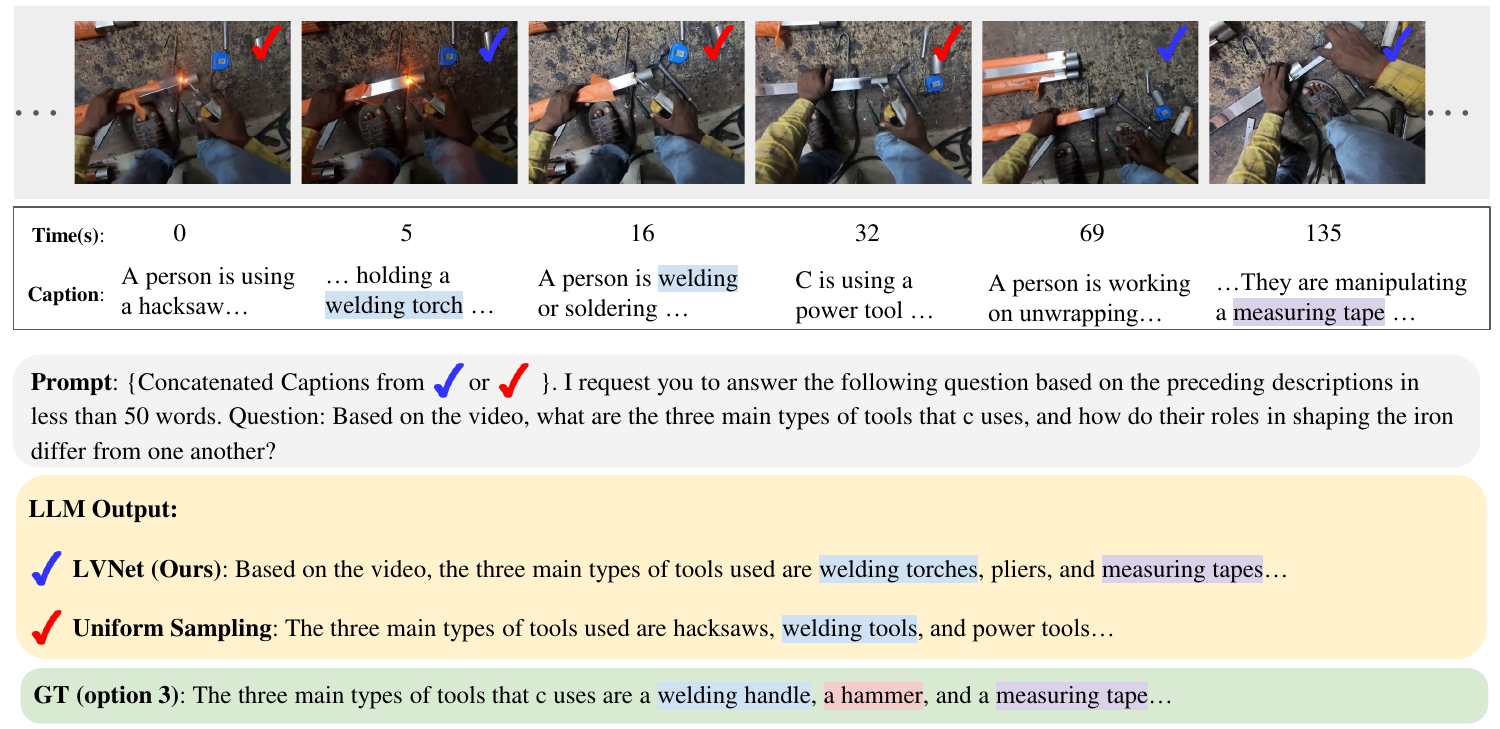}
\caption{\textbf{Open-ended Responses from LVNet vs Uniform Sampling}: The frames chosen by \modelname~and the naive uniform sampling method are indicated with blue and red checkmarks, respectively. \modelname~identifies both welding torches and measuring tapes, choosing the correct option, whereas uniform sampling only detects welding tools and selects the incorrect answer. The blue, red, and purple highlights correspond to the three main activities in the video—welding a handle, using a hammer, and using a measuring tape, respectively.
}\label{fig:qualanswer}
\end{figure*}

We compare open-ended responses of \modelname~and the uniform sampling method in \Cref{fig:qualanswer} to understand the effectiveness of the hierarchical keyframe selector in \modelname. The frames chosen by \modelname~and the naive uniform sampling method are indicated by blue and red checkmarks in the images, respectively. \modelname~selects frames at 5, 69, and 135 seconds by executing the hierarchical keyframe selector and generates captions based on those frames. When we feed the concatenated captions to the LLM to answer the given question: \textit{"Based on the video, what are the three main types of tools that C uses..."} in an open-ended manner, the output identifies two main activities: welding torches and measuring tapes, among the three main activities described in Option 3 (welding handle, hammer, measuring tape), which is the correct answer, leading \modelname~to choose the correct option. 

In contrast, the uniform sampling method selects frames at 0, 16, and 32 seconds and generates captions based on those frames. Similarly, when we feed the concatenated captions to the LLM to answer the same question, the output identifies only one activity—welding tools—resulting in the selection of the incorrect option. This example highlights the importance of keyframe selection and demonstrates the effectiveness of hierarchical keyframe selection in \modelname.

\section{Algorithms in Detail}
\label{app:hksalgorithm}
Our algorithms are presented in full detail in \Cref{alg:tsc}, \Cref{alg:ckd}, and \Cref{alg:fkd}. TSC in \Cref{alg:tsc} extracts per-frame visual features using \resneteight, followed by an iterative clustering procedure to identify $n$ non-overlapping frame sets. Within each of the $n$ sets, we uniformly sample $\leq \tau$ frames, obtaining a total of $T_a \leq \tau \times n$ frames. For example, \modelname sets $\psi = 5, \lambda = 12, \tau = 18$, resulting in approximately $n \sim 25$ and $T_a \sim 390$ on the \egoschema~dataset. CKD in \Cref{alg:ckd} selects top $L$ frames based on similarity/confidence scores, which are calculated using cosine similarity between frames and keywords with \clipb. \modelname employs $L=32, len(K) \leq 25$ on the \egoschema~dataset. FKD in \Cref{alg:fkd} sorts frames and their corresponding keywords by confidence scores, and reorder the $K$ frames with the lowest scores temporally. It groups frames sequentially into visual templates, each consisting of $N$ frames. From each template, the $M$ frames and keywords most relevant among the $N$ pairs are selected using \gpto. We set $L=32, K=16, N=8, M=3$.

\begin{figure}[t]
    \centering
    \removelatexerror
    \scalebox{0.83}{
    \begin{minipage}{0.5\textwidth}
        \centering
        \input{figures/algo_tsc}
    \end{minipage}
    }
\end{figure}

\begin{figure}[t]
    \centering
    \removelatexerror
    \scalebox{0.83}{
    \begin{minipage}{0.5\textwidth}
        \centering
        \input{figures/algo_ckd}
    \end{minipage}
    }
\end{figure}

\begin{figure}[t]
    \centering
    \removelatexerror
    \scalebox{0.83}{
    \begin{minipage}{0.5\textwidth}
        \centering
        \input{figures/algo_fkd}
    \end{minipage}
    }
\end{figure}

\section{Prompting: Fine Keyframe Detector}
\label{app:prompt_fkd}

\begin{figure*}[t]
\centering
\includegraphics[width=0.9\linewidth]{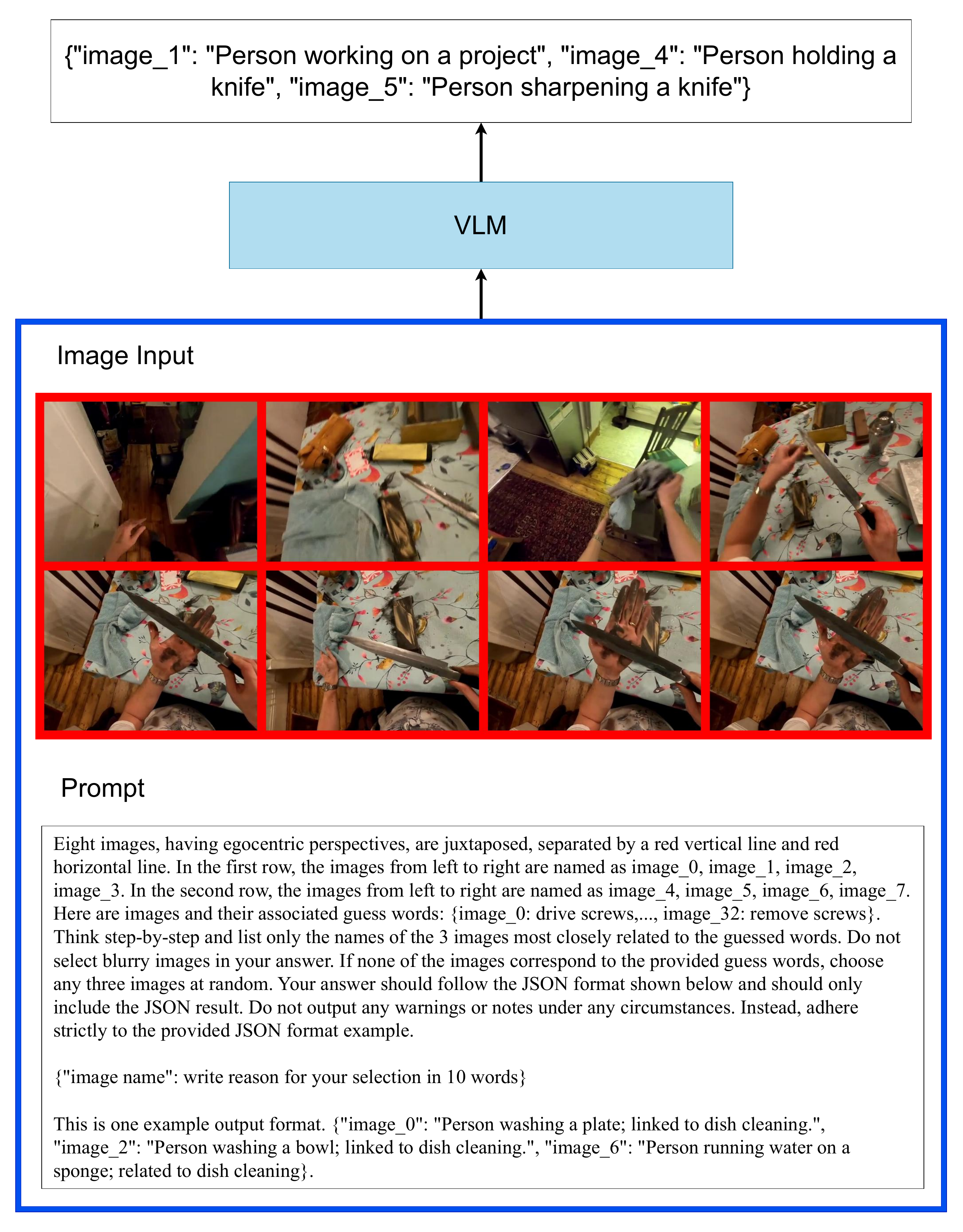}
\caption{\textbf{Prompt for Fine Keyframe Detection}: The figure illustrates the input image, the prompt provided to the VLM, and the output. The input image represents a visual template composed of eight frames, and the prompt requests the three best frames along with their corresponding keywords. The output displays the top three selected frames and their associated keywords.
}\label{fig:fkd_prompt}
\end{figure*}

We prompt the VLM to select frames that are most compatible with the list of given keywords. Each template image contains 8 images, and their order is described in language (e.g. top left to right, bottom left to right) and the VLM outputs the selected images according to our prompting as described in \Cref{fig:fkd_prompt}.

\section{Integration to Existing Methods} 
\label{app:integrations}
Our LVNet has been successfully integrated with other works for evaluation on long video benchmarks. For example, in \citet{rana2024mvu} LVNet is integrated with their proposed MVU to gain further performance boosts on the EgoSchema and NextQA benchmarks.

\section{Comparison with Other Keyframe Selection Methods} 
\label{app:other_work}
We aim to highlight the main advantage of the \hksfull over other existing keyframe selection methods. Models like \videoagent, \videotree, and \traveler~provide useful comparisons, as they utilize keyframe selection mechanism with similar or different scale of frames. \videoagent~and \traveler~rely on uniform frame selection in the first iteration without analyzing the entire video even though they perform non-uniform sampling in the next iterations. They identify important segments based solely on these initial frames and the LLM’s response, which can be problematic if the initial uniformly selected frames are not representative of the entire video or if the LLM misinterprets the captions and prompts. In such cases, the LLM might incorrectly identify segments for further analysis. If the LLM fails to pinpoint the correct segment initially, the entire process can break down because subsequent frames will be similar to the first set, leading the LLM to continuously select frames within or near the initial segment. Additionally, for videos that are as challenging or more difficult than \egoschema~in terms of temporal complexity and activities, existing keyframe selection models such as \videoagent, \videotree, and \traveler~may require numerous iterations by running heavy visual/language models to finalize keyframes selection. This results in higher computational and latency costs, as it necessitates numerous runs of resource-intensive VLM and LLM models.

In contrast, our method analyzes the entire video with high frame rates using a lightweight \resneteight~\citep{he2016resnet} and segments the video non-uniformly based on scene continuity. We then select several frames in each segment by measuring feature similarity between frame features and keywords using the \clipb~(0.12B) \citep{Ranasinghe2022PerceptualGI} which is lighted than VideoAgent’s EVA-CLIP-8Bplus (8B). By reviewing the entire video and non-uniformly selecting keyframes based on scene continuity and similarity scores, these keyframes accurately represent the question-based important frames distribution in the entire video. Furthermore, we use VLM for a fine-grained selection of keyframes, improving keyframe selection when CLIP-B/16 struggles to understand detailed atomic activities in the frames. By hierarchically segmenting the video with different modules, the resulting segments and keyframes are more reliable than those from VideoAgent. Even with more challenging videos, our process only needs to go through the video once to collect keyframes, maintaining computational efficiency.

\Cref{fig:kfscomparison4} visualizes the differences of the keyframe selection mechanism bewtween \ours~and \videoagent. On the left, \ours~begins with uniformly sampled frames and filters them through multiple stages, resulting in a non-uniform distribution of frames over time. First, the temporal scene clustering (TSC) selects some frames that represent temporally distinct activities. Next, the coarse keyframe detector (CKD) targets frames most relevant to the question. Finally, the fine keyframe detector (FKD) further refines this selection to ensure the keyframes accurately capture the activity in question. As a result, \ours~produces 12 frames, with 8 of them (67\%) directly depicting "usage of phones," which is the correct answer and leads the model to select the right option. On the right, \videoagent~also starts with the uniform frames but relies on a LLM to request additional frames. Since the initial frames do not capture enough relevant content, the LLM again selects frames uniformly, adding more irrelevant samples that lack the crucial information about "usage of phones." As a result, \videoagent~ultimately selects the wrong option.
\label{sec:method:comparison}

\end{document}